\documentclass[10pt,twocolumn,letterpaper]{article}

\usepackage{iccv}
\usepackage{times}
\usepackage{epsfig}
\usepackage{graphicx}
\usepackage{amsmath}
\usepackage{amssymb}

\usepackage{mathtools}
\usepackage{overpic}
\usepackage{microtype}
\usepackage{paralist}
\usepackage{placeins}
\usepackage{booktabs}
\usepackage{bbding}
\usepackage{comment}
\usepackage{subcaption}
\usepackage[normalem]{ulem}
\usepackage{multicol}
\usepackage{wrapfig}
\usepackage{makecell}
\usepackage{rotating}
\usepackage{physics}
\usepackage{titling}
\usepackage[resetlabels,labeled]{multibib}
\newcites{A}{Additional References}

\usepackage{color}
\definecolor{turquoise}{cmyk}{0.65,0,0.1,0.3}
\definecolor{purple}{rgb}{0.65,0,0.65}
\definecolor{dark_green}{rgb}{0, 0.5, 0}
\definecolor{orange}{rgb}{0.8, 0.6, 0.2}
\definecolor{red}{rgb}{0.8, 0.2, 0.2}
\definecolor{darkred}{rgb}{0.6, 0.1, 0.05}
\definecolor{blueish}{rgb}{0.0, 0.3, .6}
\definecolor{light_gray}{rgb}{0.7, 0.7, .7}
\definecolor{pink}{rgb}{1, 0, 1}
\definecolor{greyblue}{rgb}{0.25, 0.25, 1}

\usepackage{blindtext}

\newcommand{\loss}[1]{\mathcal{L}_\text{#1}}

\renewcommand{\real}{\mathbb{R}}

\newcommand{\outrdnc}{L_o}
\newcommand{\inrdnc}{L_i}
\newcommand{\shbase}{Y}
\newcommand{\vis}{V}
\newcommand{\normatten}{H}
\newcommand{\viewangle}{\boldsymbol{\omega}}
\newcommand{\inview}{\viewangle_i}
\newcommand{\outview}{\viewangle_o}
\newcommand{\pnt}{\mathbf{x}}
\newcommand{\diffuse}{d}
\newcommand{\specular}{s}
\newcommand{\brdf}{\rho}
\newcommand{\normal}{\mathbf{n}}
\newcommand{\lightsh}{l}
\newcommand{\prt}{t}

\newcommand{\rfview}{\viewangle_r}
\newcommand{\shdeg}{N}

\newcommand{\spectint}{K}

\newcommand{\feature}{\mathbf{F}}

\newcommand{\ray}{\mathbf{r}}
\newcommand{\raylen}{\tau}
\newcommand{\mcolor}{\mathbf{C}}
\newcommand{\normalmap}{\mathbf{N}}
\newcommand{\mask}{\mathbf{M}}

\DeclareMathOperator{\Lap}{D_{\textrm{xy}}}

\makeatletter
\DeclareRobustCommand\onedot{\futurelet\@let@token\@onedot}
\def\@onedot{\ifx\@let@token.\else.\null\fi\xspace}

\def\eg{\emph{e.g}\onedot} 
\def\ie{\emph{i.e}\onedot}

\makeatother
\usepackage{multirow}

\usepackage{array}
\newcolumntype{L}[1]{>{\raggedright\let\newline\\\arraybackslash\hspace{0pt}}m{#1}}
\newcolumntype{C}[1]{>{\centering\let\newline\\\arraybackslash\hspace{0pt}}m{#1}}
\newcolumntype{R}[1]{>{\raggedleft\let\newline\\\arraybackslash\hspace{0pt}}m{#1}}

\newcommand{\moniker}{LumiGAN}

\usepackage[pagebackref=true,breaklinks=true,letterpaper=true,colorlinks,bookmarks=false]{hyperref}

\usepackage[capitalize]{cleveref}
\crefname{section}{Sec.}{Secs.}
\Crefname{section}{Section}{Sections}
\Crefname{table}{Table}{Tables}
\crefname{table}{Tab.}{Tabs.}
\Crefname{figure}{Figure}{Figures}
\crefname{figure}{Fig.}{Figs.}
\creflabelformat{equation}{#2#1#3}

\iccvfinalcopy 

\def\iccvPaperID{5278} 

\ificcvfinal\fi

\begin{document}
\title{LumiGAN: Unconditional Generation of Relightable 3D Human Faces}

\author{
Boyang Deng
\qquad
Yifan Wang
\qquad
Gordon Wetzstein\\[0.1in]
Stanford University
}
\date{}

\maketitle
\ificcvfinal\thispagestyle{empty}\fi

\begin{abstract}
Unsupervised learning of 3D human faces from unstructured 2D image data is an active research area. While recent works have achieved an impressive level of photorealism, they commonly lack control of lighting, which prevents the generated assets from being deployed in novel environments. To this end, we introduce \moniker{}, an unconditional Generative Adversarial Network (GAN) for 3D human faces with a physically based lighting module that enables relighting under novel illumination at inference time. Unlike prior work, \moniker{} can create realistic shadow effects using an efficient visibility formulation that is learned in a self-supervised manner.
\moniker{} generates plausible physical properties for relightable faces, including surface normals, diffuse albedo, and specular tint without any ground truth data. In addition to relightability, we demonstrate significantly improved geometry generation compared to state-of-the-art non-relightable 3D GANs 
and notably better photorealism than existing relightable GANs.
\end{abstract}

\newcommand\blfootnote[1]{%
  \begingroup
  \renewcommand\thefootnote{}\footnote{#1}%
  \addtocounter{footnote}{-1}%
  \endgroup
}

\blfootnote{Project page: \href{https://boyangdeng.com/projects/lumigan}{\nolinkurl{boyangdeng.com/projects/lumigan}}.}

\section{Introduction}\label{sec:intro}
\begin{figure}[htbp]
    \centering
    \includegraphics[width=\linewidth]{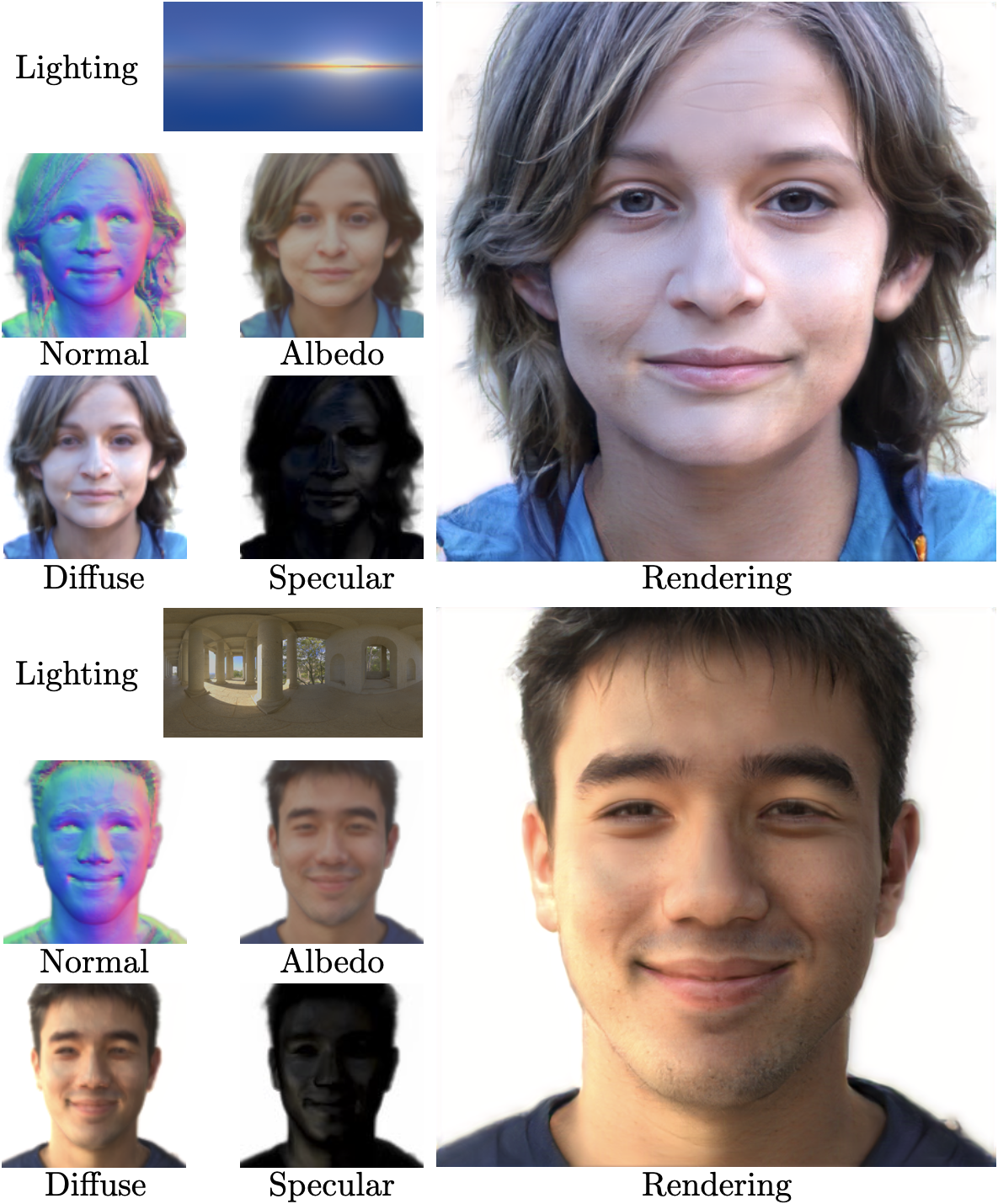}
    \caption{
    \textbf{\moniker{}} is an unconditional 3D GAN for photorealistic relightable 3D faces, trained solely from unstructured single-view images under unknown and varying lighting conditions. Using a physically based lighting module and self-supervised training, it generates relevant physical properties (left) from a single latent code, allowing the generated 3D face to be relit under novel illuminations (right) with physically plausible shadow effects.
    }
    \label{fig:teaser}
\end{figure}
Synthetic 3D digital humans are valuable assets for a wide variety of applications, ranging from virtual or augmented reality and visual effects in film and gaming to data generation for training visual perception systems.
By combining Generative Adversarial Networks~\cite{goodfellow2014generative,karras2019style} (GANs) and 
neural rendering~\cite{tewari2022advances}, several recent methods have demonstrated unconditional generation of photorealistic 3D human faces (e.g.,~\cite{chan2022efficient}).
Existing 3D GANs, however, pay little attention to making the generated assets editable, for example prohibiting digital humans to be deployed in new lighting conditions, \ie lacking relightability.
%

Foundational research in the graphics community has enabled photorealistic quality for relightable humans acquired with specialized capture rigs~\cite{debevec2000acquiring}. More recent works have extended these methods using neural rendering~\cite{zhang2021nerfactor,zhang2021physg,lyu2022nrtf, verbin2022ref}. While all these approaches can achieve impressive results for reconstructing individuals captured with the required multi-camera rigs, it is challenging to scale them to generative settings due to the limited amount and diversity of available multi-view data and their computationally costly light transport models.

Very recent efforts have developed 3D GAN architectures for unconditional generation of relightable 3D human faces~\cite{pan2021shading,tan2022voluxgan}.
Trained with unstructured, single-view images of humans in fixed and unknown lighting conditions, these methods learn to generate shape, albedo, normals, specular material properties, or other components that are used by a physically inspired lighting model to compute the final color. Training a 3D GAN, however, is computationally expensive, requiring millions of neural rendering passes. For this reason, existing 3D GANs use overly simplified lighting models that are not physically accurate. This approach inevitably limits the expressiveness of the generator, which has to be compensated by post-rendering 2D filters in a non-physical manner, causing view inconsistency and artifacts.

We introduce \moniker{}, a 3D GAN framework enabling unconditional generation of relightable human faces with state-of-the-art visual quality and view consistency.
Our framework uses an expressive, yet efficient physically based lighting model to learn to generate geometry, albedo, specular tint, and visibility components of a person's face in an unsupervised manner from large image datasets.
While prior works either ignore visibility or require casting costly secondary rays,
our carefully designed visibility term is predicted, thus computationally efficient and feasible in a generative setting, but also self-supervised to be physically plausible after training.
Due to its effectiveness in modeling shadows and self-occlusion in a physically accurate manner, our visibility term, among other important insights, enables significantly improved image quality and consistency for changing camera poses and illumination.
Similar effects could only be hallucinated in prior work, at the cost of degraded multi-view and illumination consistency.

Our key technical contributions include:
\begin{compactitem}
\item an expressive yet efficient lighting model for neural rendering, which accounts for self-occlusion to faithfully model complex self-shadowing;
\item a self-supervised framework that learns to generate disentangled geometry, albedo, specular tint, normals, and visibility from large-scale casual image datasets without controlled lighting or ground truth supervision.
\end{compactitem}

Our experiments demonstrate that the proposed generative model can produce relightable 3D human faces with state-of-the-art photorealism, outperforming existing relightable 3D GANs by a large margin.

Our code and pre-trained models will be published upon acceptance.

\section{Related Work}\label{sec:rw}
Our work is inspired by numerous prior works in graphics and computer vision, which we briefly summarize here.

\paragraph{\bf Relighting from Studio Captures.}
Previous work tackled the relighting problem using precisely calibrated and controllable studios~\cite{debevec2000acquiring, beeler2010high, collet2015high, joo2015panoptic}.
One of the most important works in this field is the light stage~\cite{debevec2000acquiring}, which was the first system to create a highly accurate model of a subject with controllable illumination that could be lit in various realistic ways.
Over the last two decades, light stages have advanced to account for things like changing facial expressions~\cite{hawkins2004animatable}, complex light reproduction~\cite{debevec2002lighting}, full-body support~\cite{chabert2006relighting}, and polarized spherical gradient illumination~\cite{ma2007rapidacquisition, ghosh2011multiview}.
This technology is now a foundation of modern film production~\cite{debevec2012light}. More recently, neural networks have been used to improve the light stage processing pipeline~\cite{meka2019deep,meka2020deep,zhang2021neural,bi2021deep,zhang2021nerfactor,zhang2021physg,sengupta2021light}.
Neural radiance fields (NeRFs)~\cite{mildenhall2020nerf} have also been applied to light stage data for novel view synthesis and relighting~\cite{srinivasan2021nerv}. Other works have trained deep neural networks using light stage data to enable relighting for less-constrained conditions during test time, for example using a single input image~\cite{sun2019single,pandey2021total,yeh2022learning} or monocular video~\cite{zhang2021neuralvideo}.
However, light stages and other capture studios are expensive and difficult to calibrate and maintain. This makes it challenging to capture a large diversity of subjects, and difficult to use them to train data-driven algorithms.
\moniker{}, in contrast, uses unstructured image collections of single-view portraits in uncontrolled lighting conditions.
Such data is accessible to everyone free of charge and presents richer diversity.

\paragraph{\bf Relighting from Casual Image Collections.}
Relighting a subject given limited images with little to no knowledge of the illumination conditions requires decomposing the observations into several components, including geometry, material properties, and lighting, which is an ill-posed problem.
Throughout the last decades the research community tackled this ambiguity predominantly by introducing hand-crafted priors, linking to the early study of intrinsic images~\cite{barrow1978recovering, land1971lightness}.
To date, most relighting works~\cite{wang2022sunstage, zhang2021nerfactor, lyu2022nrtf, chen2022relighting4d, boss2021nerd, boss2021neuralpil} adopt variants of the prior introduced by Barron and Malik~\cite{barron2014shape}, which follows smoothness and parsimony principles.
Our method also adopts a simple smoothness prior to enhance the decomposition quality, but we do so in a generative framework and only for the albedo prediction.
The decomposition of geometry, material, and lighting is largely enabled by the inherent physical constraints imposed by our physically based lighting formulation.
Similarly, our generative framework allows us to learn the natural distribution of geometry and material from diverse training data, which leads to more realistic relighting results under real world illumination conditions.

\paragraph{\bf Editable 3D Human Generation.}
The task of generating 3D digital humans has recently made tremendous progress~\cite{nguyen2019hologan,schwarz2020graf,Niemeyer2020GIRAFFE,chan2021pi,gu2021stylenerf,chan2022efficient,skorokhodov2022epigraf,or2022stylesdf,xu20223d,zhou2021cips,xue2022giraffehd,zhang2022mvcgan,deng2022gram,xiang2022gramhd} by extending 2D GANs~\cite{goodfellow2014generative,karras2019style,karras2020analyzing} to 3D with the help of neural rendering~\cite{tewari2022advances}.
Recent work sought to inject controllability and post-generation editability into 3D GANs. This can be achieved by training generative networks in a supervised manner using labeled attributes such as lighting or facial expressions~\cite{hong2022headnerf,liu20223d}. Alternatively, unsupervised approaches have successfully generated articulated digital humans by including physically based body pose deformation modules into the generator~\cite{bergman2022generative,noguchi2022unsupervised,zhang2022avatargen}.

Our 3D GAN uses an unsupervised learning approach that augments the generator network using a physically based lighting module with the goal of learning to generate physically plausible albedos, diffuse and specular components, among other parameters.
The ideas explored by ShadeGAN~\cite{pan2021shading} are related in that they decompose the RGB color into ambient and diffuse, and thus enabled rudimentary relighting using a simple Lambertian lighting model.
Closest to our work is VoluxGAN~\cite{tan2022voluxgan}, which directly uses preconvolved light maps, similar to~\cite{pandey2021total}, to compute diffuse and specular shadings, thereby achieving more realistic relighting with a Phong lighting model~\cite{phong1975illumination}.
Yet, neither of these lighting models are expressive enough to capture complex shadows or specularities, instead these effects are ``sneaked in'' by the post-rendering 2D convolutions, which leads to compromised view consistency and decomposition quality.
Consequently, VoluxGAN requires generating reference albedo and normals for supervision, which may contain biases and are in general difficult to obtain.
In comparison, we leverage a physically more accurate yet efficient lighting model that uses learned geometry, visibility and various material properties to represent the diffuse and specular shadings, accounting for high-frequency illumination effects and shadows.
This not only leads to significant improvement in visual quality, but also better geometry details thanks to the stronger leverage of geometric cues in the lighting model.

\begin{figure*}[htbp]
    \centering
    \includegraphics[width=\textwidth]{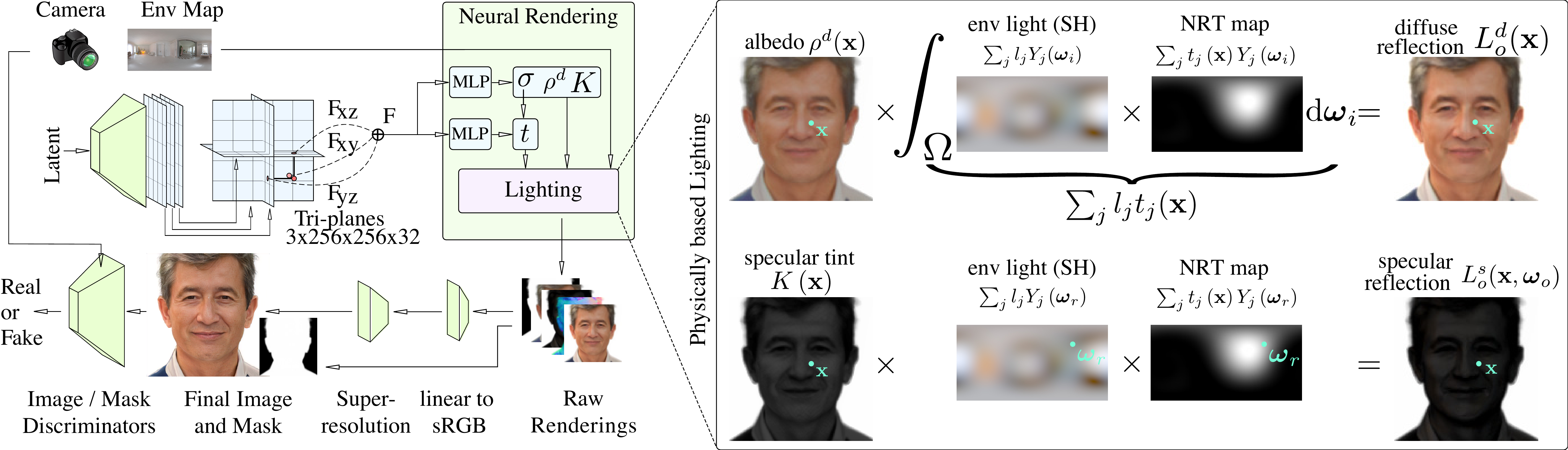}
    \caption{\textbf{Method Overview.} \moniker{} integrates a physically based lighting module to a $3$D GAN to generate relightable 3D faces.
    The StyleGAN-inspired generator generates triplane features from a latent code, which are then mapped to volume density \(\sigma\), diffuse albedo \(\brdf^{d}\), specular tint \(\spectint\), and visibility \(\left[ \prt_{1}, \dots, \prt^{\shdeg}\right]\).
    Given an environment map projected to spherical harmonics coefficients \(\left[ \lightsh_{1}, \dots, \lightsh_{\shdeg} \right]\), the physically based lighting module uses these generated properties to render the lit image, particularly accounting for physically plausible shadows.
    The rendered lit image, transformed from linear color space to sRGB, is passed into a super-resolution block with other features to produce the final high-resolution image.}
    \label{fig:pipeline}
\end{figure*}
\section{Method}\label{sec:method}
Core to \moniker{} is an expressive while scalable physically based lighting module.
In this section, we first derive the mathematical formulation of this module (\cref{sec:lighting}), then introduce the corresponding neural representations with a focus on visibility modeling through our novel Neural Radiance Transfer (\cref{sec:autoprt}), and finally present the integration of this formulation to $3$D GANs (\cref{sec:gan}).

\subsection{Physically Based Lighting}
\label{sec:lighting}

Relighting frameworks often build on a variant of the rendering equation~\cite{kajiya1986rendering} that assumes that direct illumination from distant sources dominates the incident radiance at a point.
This implies that the outgoing radiance \(\outrdnc(\pnt, \outview)\) at a point \(\pnt\) in the viewing direction \(\outview\) is represented by the following integral over a sphere
\begin{equation}
\scalebox{0.86}{$\outrdnc(\pnt, \outview) = \int_{\Omega} \inrdnc(\inview)\brdf\left( \pnt, \inview, \outview \right) \vis(\pnt, \inview) \normatten(\pnt, \inview) \dd{\inview},$}
\label{eq:general_lighting}
\end{equation}
where \(\inrdnc\left( \inview \right)\) denotes the incident distant light arriving from direction \(\inview\);
$\brdf$ refers to the bidirectional reflection distribution function (BRDF);
$\vis$ indicates the visibility of incident direction $\inview$ at $\pnt$;
$\normatten\left( \pnt, \inview \right) = \max(0, \normal( \pnt ) \cdot \inview)$ is the attenuation of incident radiance based on the normal $\normal(\pnt)$ at $\pnt$ and the incident direction $\inview$, clamped to only the upper hemisphere.
Here, we assume the point $\pnt$ is not an emitter but only reflects light.
Following common practice~\cite{zhang2021nerfactor, wang2022sunstage, srinivasan2021nerv}, we split the integral in~\cref{eq:general_lighting} into a diffuse integral and a specular integral
\begin{equation}
\begin{aligned}
&\outrdnc(\pnt, \outview) = \underbrace{\frac{\brdf^{\diffuse}(\pnt)}{\pi} \!\! \int_{\Omega} \!\! \inrdnc(\inview) \vis(\pnt, \inview) \normatten(\pnt, \inview) d{\inview}}_{\textstyle \textrm{Diffuse}, \outrdnc^{\diffuse}(x)} \\
&+ \underbrace{\int_{\Omega} \inrdnc(\inview)\brdf^{\specular}\left( \pnt, \inview, \outview \right) \vis(\pnt, \inview) \normatten(\pnt, \inview) \dd{\inview}}_{\textstyle \textrm{Specular}, \outrdnc^{\specular}(x, \outview)}
\end{aligned}
\label{eq:breakdown_lighting}
\end{equation}
where $\brdf^{\diffuse}$ is the spatially-varing diffuse albedo, $\brdf^{\specular}$ the spatially-varying specular BRDF, and $\pi$ is a normalization term for the conservation of energy.

Evaluating the integral in \cref{eq:breakdown_lighting} is costly and computationally prohibitive for GAN training,
where millions of images need to be rendered. One can conclude from inspecting the diffuse part of \cref{eq:breakdown_lighting} that the efficiency bottleneck is two-fold:
(1) Computing the visibility $\vis$ requires casting secondary visibility rays;
(2) The numerical estimation of the spherical integral requires sampling incident directions
from the environment maps.

Our solution to the aforementioned challenges takes inspiration from Precomputed Radiance Transfer~\cite{sloan2002precomputed} (PRT), a technique commonly used in real-time graphics engines.
Specifically, we approximate the compound of $\vis$ and $\normatten$ using the first $\shdeg$ spherical harmonics (SH) bases:
\begin{equation}
    \vis(\pnt, \inview)\normatten(\pnt, \inview) \approx \sum_{j=1}^{\shdeg} \prt_j(\pnt) \shbase_j(\inview).
\label{eq:prt_proj}
\end{equation}
Therefore, we can rewrite the diffuse part in \cref{eq:breakdown_lighting} as follows:
\begin{align}
    \outrdnc^{\diffuse}(\pnt) &\approx \frac{\brdf^{\diffuse}(\pnt)}{\pi} \!\! \int_{\Omega} \!\! \inrdnc(\inview) \sum_{j=1}^{\shdeg} \prt_j(\pnt) \shbase_j(\inview) d{\inview} \\
    &= \frac{\brdf^{\diffuse}(\pnt)}{\pi} \!\! \sum_{j=1}^{\shdeg} \prt_j(\pnt) \int_{\Omega} \!\! \inrdnc(\inview) \shbase_j(\inview) d{\inview} \label{eq:diffuse_second_final}\\
    &= \frac{\brdf^{\diffuse}(\pnt)}{\pi} \!\! \sum_{j=1}^{\shdeg} \prt_j(\pnt) \lightsh_j \label{eq:diffuse_final}
\end{align}
where \cref{eq:diffuse_final} is based on the observation that the integral in \cref{eq:diffuse_second_final} is effectively projecting the illumination to the $j$-th spherical harmonics basis.
Using this formulation, we can effectively evaluate the light source visibility weighted by the normal attenuation at any point $\pnt$ by directly querying the function $\prt$ instead of casting visibility rays;
at the same time it also reduces the integral to a dot product whose computation complexity is constant to the resolution of area light resource we use for illumination.
Different from the PRT formulation~\cite{sloan2002precomputed}, which includes the albedo in the precomputation and uses a per color-channel radiance transfer for interreflection, we simplify the formulation to a single-channel $\prt$ involving $\vis$ and $\normatten$, based on the observation that interreflection is generally insignificant on human faces.
By isolating albedo from PRT, we also obatain better controllability over the model for one can edit the standalone albedo.

As for the specular part, we observe that high frequency and concentrated highlights are rare in common objects that are mostly diffuse, \eg human faces.
Hence, we mainly focus on modeling low-frequency specular reflections.
This leads us to reuse the band-limited spherical harmonics projection of illumination we have calculated for diffuse reflections.
Moreover, because the illumination is already low-pass filtered, we reduce the specular reflection to a single perfect reflection direction $\rfview$.
We argue that this is approximately equivalent to simulating a rough specular surface, such as human skin, with high-frequency illumination.

Thus, we rewrite the specular integral in \cref{eq:breakdown_lighting} as:
\begin{equation}
\outrdnc^{\specular}(\pnt, \outview) = \spectint(\pnt) \underbrace{\sum_{j=1}^{\shdeg}\lightsh_j \shbase_j(\rfview)}_{L_i \left( \rfview \right)} \sum_{j=1}^{\shdeg}\prt_j(\pnt) \shbase_j(\rfview),
\label{eq:specular}
\end{equation}
where the reflection direction depends on view angle and the surface normal \(\rfview = 2(\outview \cdot \normal(x)) \normal(x) - \outview\).
Combining \cref{eq:diffuse_final} and \cref{eq:specular}, we can now efficiently compute the reflection of lights from area light sources.

\subsection{Neural Radiance Transport}
\label{sec:autoprt}
Traditionally in real-time graphics, given an object with fixed geometry and appearance, radiance transfer is precomputed and cached beforehand and reused for an object under various illumination.
However, such conditions do not hold in generative frameworks as we usually only know the actual geometry and appearance at the time of rendering, \ie at inference.
Instead, we propose to \emph{predict} the radiance transfer coefficients.
Specifically, based on \cref{eq:breakdown_lighting,eq:diffuse_final,eq:specular}, our model predicts for each point $\pnt$:
\begin{inparaenum}[(1)]
\item $\brdf^{\diffuse}(\pnt)\in \real^{3}$, spatially-varying albedo;
\item $[\prt_1(\pnt), \dots, \prt_{\shdeg}(\pnt)]\in \real^{N}$, spatially-varying radiance transfer coefficient;
\item $\spectint(\pnt)\in\real$, spatially-varying specular tint.
\end{inparaenum}
We assume that the spherical harmonics projections of the illumination $[\lightsh_1, \dots, \lightsh_{\shdeg}]$ are given.

Meanwhile, we design our $3$D representation as neural fields to be compatible to volumetric rendering in the following form:
\begin{equation}
\begin{aligned}
\mcolor\left(\ray,\outview \right) &= \int_{\raylen_{n}}^{\raylen_{f}} T\left( \raylen \right)\sigma\left( \ray\left( \raylen \right) \right) \outrdnc(\pnt, \outview) \dd \raylen\\
T\left( \raylen \right) &= \exp\left( -\int_{\raylen_{n}}^{\raylen} \sigma\left( \ray\left( s \right) \right) \dd s\right),
\end{aligned}
\label{eq:volum_render}
\end{equation}
where we define a ray \(\ray = \vb{o} + \raylen\vb{d}\) with \( \raylen\in\left[ \raylen_{n}, \raylen_{f} \right]\) and $\vb{d}=-\outview$, and $\sigma(x)$ the volume density function that our model has to predict as well.
Note that to compute the perfect reflection $\rfview$ in \cref{eq:specular}, we need the normal at the lighting point $x$.
With the volume density formulation, we can compute the normal $\normal(x)$:
\begin{equation}
    \normal(x) = - \frac{\nabla \sigma(x)}{|| \nabla \sigma(x) ||}.\label{eq:normal}
\end{equation}

While we have formulated the predictions of all quantities for lighting, without any constraints, the model is not able to decompose each quantity in a physically correct way thus unable to produce plausible results under novel illuminations.

To alleviate this issue, we propose to self-supervise the predicted PRT coefficients by enforcing their physical interdependence with the density term.
Specifically, \(\vis\left( \pnt, \inview \right)\normatten\left( \pnt, \inview \right)\) can alternatively be computed from the density field, where the visibility is equivalent to the transmittance of the ray from the point $\pnt$ to a virtual camera positioned along the incident ray:
\begin{equation}
  \vis\left( \pnt, \inview \right) = \exp\left( -\int_{\raylen_{n}}^{\raylen_{\pnt}} \sigma\left( \ray\left( s \right) \right) \dd s\right).\label{eq:gt_vis}
\end{equation}
Conveniently, for \(\inview=\outview\) the density values are readily available from the volume rendering in \cref{eq:volum_render}.
Consequently, we can supervise $[\prt_1(\pnt), \dots, \prt_{\shdeg}(\pnt)]$ by
\begin{equation}
    \scalebox{0.9}{$\loss{nrt}(\pnt) = \left( \sum_{j}^{\shdeg}\prt_j\left( \pnt \right) Y_j\left( \outview \right) - \vis\left( \pnt, \outview \right)\normatten\left( \pnt, \outview \right) \right)^2.$}
\label{eq:prt_loss}
\end{equation}
In practice, we further enhance this loss by randomly sampling few extra rays to handle directions not covered by the camera poses, \eg top view and bottom view.

\subsection{Relightable 3D GAN}
\label{sec:gan}
The backbone of LumiGAN closely follows EG3D~\cite{chan2022efficient}.
As shown in \cref{fig:pipeline}, the neural renderer outputs a set of raw renderings at low resolution, including albedo, normal map, diffuse, specular shading, and the composed lit image.
These are fed into a super-resolution module to generate the final high-resolution lit image.
In addition, we propose the following changes to the EG3D framework to train \moniker{} from unlabeled casual image datasets such as FFHQ~\cite{karras2019style}.

While the relighting of foreground objects is well-defined, the relighting of the generated background is unclear.
Particularly, the background of real-world images can be distant and/or blurred depending on the context where the picture is taken.
Such properties are propagated to generated images via GAN training, leading to ill-defined background relighting.
We circumvent this obstacle by encouraging the 3D generative model to generate background-free 3D faces.
Specifically, we use an auxiliary foreground mask discriminator that takes in both the foreground masks of real images predicted by an off-the-shelf foreground segmentation model~\cite{modnet2022ke} and the accumulated foreground masks of generated neural fields.
With this extra discriminator, the generative model learns to clear the background instead of predicting solid background geometry as in EG3D.
Besides the added discriminator, we also modify the other discriminators by feeding them real images with masked backgrounds.
In practice, we find that using random colors as the background mask color works the best, such that the foreground can be completely solid --- imagine how a constant white background color could be used by holes in the foreground model to fake a white highlight.

\moniker{} can generate images under any illumination condition by inputting the spherical harmonics projections of the environmental lighting.
To this end, we collect a dataset of real-world high-resolution environment maps from online public sources~\cite{polyheaven} and project them to spherical harmonics bases.
During training, we randomly sample from these maps and feed them to the neural renderer to generate lit images.
We find that by using real world illuminations, our model learns to generate realistic and physically plausible geometry and reflection.

Besides the above main changes, we also apply a simple regularization to encourage spatially-smooth albedo prediction as NeRFactor~\cite{zhang2021nerfactor}, which penalizes the albedo variance of closeby points, \ie,
\begin{equation}
\resizebox{0.88\hsize}{!}{
    \(
    \loss{smooth}(\pnt) = \|\brdf^{d}\left( \pnt \right) - \brdf^{d}\left( \pnt + \boldsymbol{\epsilon}\right)\|_{1}, \boldsymbol{\epsilon}\sim\mathcal{N}\left( \mathbf{0}, 0.03\mathbf{I} \right).\label{eq:albedo_loss}
\)
}
\end{equation}

\begin{figure}[t]
    \centering
    \includegraphics[width=0.9\linewidth]{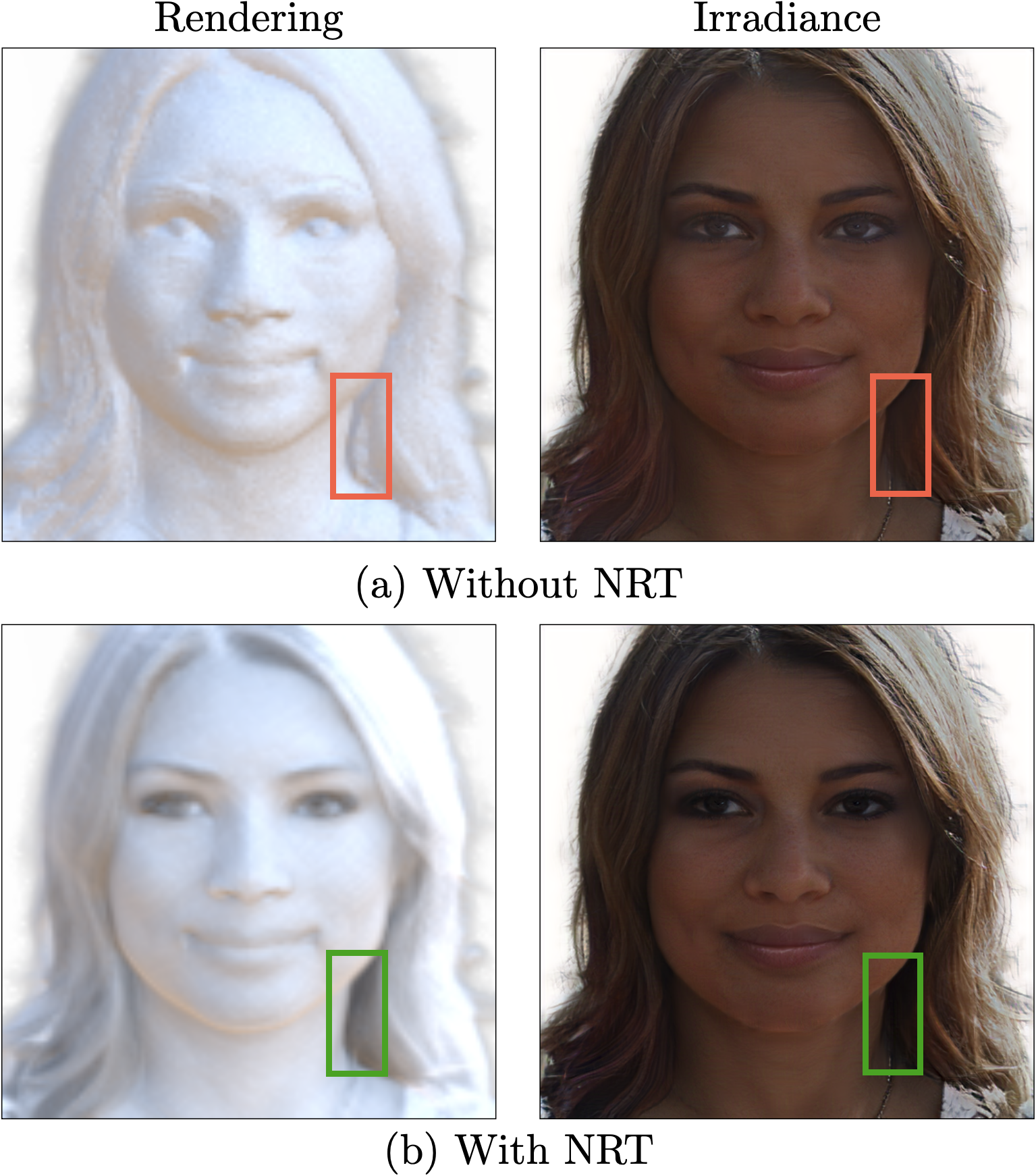}
    \caption{
    \textbf{Effect of Neural Radiance Transfer.}
    Without the proposed NRT, one can still use surface normals alone to compute the intensity of incident illumination, \ie irradiance.
    Consequently, the irradiance doesn't consider light visibility and self-shadowing (a).
    Our NRT term learns to account for visibility efficiently in a self-supervised manner, making the lighting calculations more physically correct (b).
    }
    \label{fig:w_wo_nrt}
\end{figure}

\section{Experiments}\label{sec:experiments}
\begin{table*}[htbp]
\centering\begin{tabular}{ccc*{2}{c}*{4}{c}c}
\toprule
\multirow{2}{*}{method, resolution} & \multirow{2}{*}{FID $\downarrow$} & \multirow{2}{*}{KID $\downarrow$} & \multicolumn{2}{c}{Normal} & \multicolumn{4}{c}{Id. Similarity (view) $\uparrow$} & \multirowcell{2}{Id. Similarity\\(lighting) $\uparrow$}\\\cmidrule(lr){4-5}\cmidrule(lr){6-9}
& & & Cos $\uparrow$ & LapL1 $\downarrow$ & -0.5 & -0.25 & 0.25 & 0.5 \\\midrule
VoluxGAN, \(256^{2}\) & 59.79 & 4.124 & \underline{0.78} & \textbf{0.033} & 0.606 & 0.774 & 0.800 & 0.599 &  0.831 \\
EpiGRAF, \(512^{2}\) & 9.92 & 0.453 & 0.75 & 0.098 & \underline{0.756} & \textbf{0.911} & \textbf{0.910} & \underline{0.735} & -   \\
EG3D, \(512^{2}\) & \underline{4.70} & \textbf{0.132} & 0.73 & 0.088 & 0.743 & 0.814 & 0.812 & 0.741 & - \\
EG3D-noBG, \(512^{2}\) & \textbf{4.59} & \underline{0.200} & 0.75 & 0.084 & 0.684 & 0.782 & 0.795 & 0.704 & -  \\
\moniker{} (ours), \(512^{2}\) & {5.28} & {0.251} & \textbf{0.79} & \underline{0.060} & \textbf{0.765} & \underline{0.888} &  \underline{0.883} & \textbf{0.772} & \textbf{0.947}
\\\bottomrule
    \end{tabular}
    \caption{\textbf{Quantitative Comparison.}
    We compare our \moniker{} with state-of-the-art 3D relightable face generator VoluxGAN~\cite{tan2022voluxgan} and non-relightable 3D face generators, EG3D~\cite{chan2022efficient} and EpiGRAF~\cite{skorokhodov2022epigraf}, as well as a backgroud-less EG3D variant (EG3D-noBG).
    We evaluate the
    (1) image quality using FID and KID (\(\times100\));
    (2) the geometry quality using cosine similarity and the Laplacian L1 distance between the generated normal maps with reference normal maps~\cite{Abrevaya_2020_CVPR};
    (3) the identity similarity of 100 generated people under different yaw angles;
    (4) the relighting quality using the identity similarity score of 100 people relit with a pair of randomly sampled environment maps.
    Except for EG3D and EpiGRAF, the generated images are background-less, and the reference images are filtered correspondingly with a pretrained matting method~\cite{modnet2022ke}.
    }
    \label{tbl:FID}
\end{table*}
\subsection{Implementation Details}\label{sec:implementation_details}
We train our model using FFHQ~\cite{karras2019style}, a real-world human face dataset.
We follow the data preprocessing protocol of EG3D~\cite{chan2022efficient} to align the face area and estimate the camera pose.
Additionally, we generate ground truth masks using MODNet~\cite{modnet2022ke} for the foreground discriminator (\cref{sec:gan}).
We initialise our models using publicly available EG3D checkpoints to accelerate convergence.
The volumetric renderer adopts a raw resolution of \(64\times64\), and the outputs are upsampled to \(512\times512\) using the super-resolution module.
The models are trained on 8 NVIDIA RTX A6000 GPUs for $25000$ iterations, using a batch size of 48.
We set the weight of $\loss{nrt}$ to $50$ and all other loss weights to $1$.
Please see our \emph{Supplementary Material} for further implementation details.

\subsection{Relightability}
\label{sec:generation}
\paragraph{Decomposition.}
\moniker{} predicts density, visibility, diffuse albedo, and the specular tint jointly.
The proposed self-supervision, \(\loss{nrt}\), and the smooth albedo regularization, \(\loss{smooth}\), ensure\textbf{} that the decomposition is physically plausible even without ground truth supervision on any of the intermediate predictions.
Some examples of the decomposition results are visualized in \cref{fig:teaser}.

\paragraph{Shadow Effects.}
One of our key contributions is the efficient and geometry-consistent shadow thanks to the NRT term.
In \cref{fig:w_wo_nrt}, we visualise the lit image and irradiance when the NRT term (and correspondingly physically based shadows) is removed.
We can see without NRT, the face appears flatter and reveals unrealistic artifacts shown in the highlighted area, especially visible in irradiance.
Such artifacts can be detrimental to the user experience in applications, where photorealism is of paramount importance.

\begin{figure*}[htbp]
\setlength{\rotheadsize}{0.1\linewidth}
\renewcommand{\arraystretch}{0}%
\setlength\tabcolsep{0pt}
\setlength{\defaultaddspace}{0.3em}
\centering
\begin{tabular}{C{0.03\linewidth}C{0.32\linewidth}C{0.32\linewidth}C{0.32\linewidth}}
\multirow{2}{*}{\rotatebox[origin=c]{90}{ShadeGAN}}&
\includegraphics[width=\linewidth]{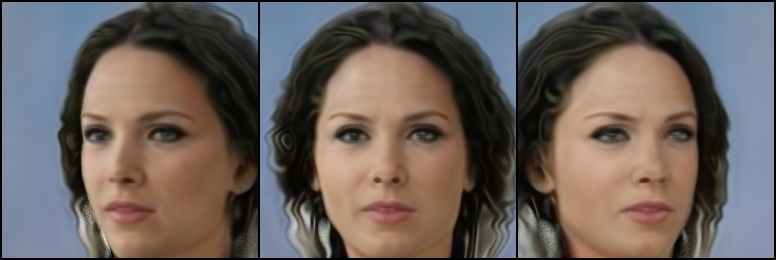} &
\includegraphics[width=\linewidth]{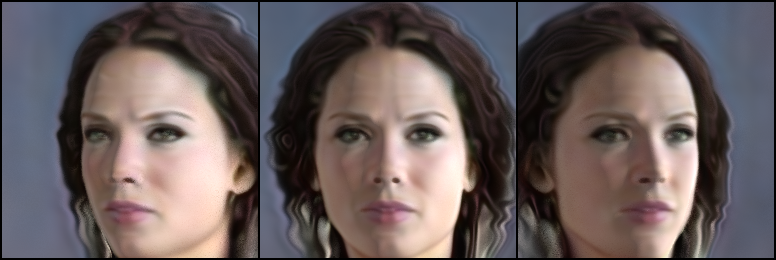} &
\includegraphics[width=\linewidth]{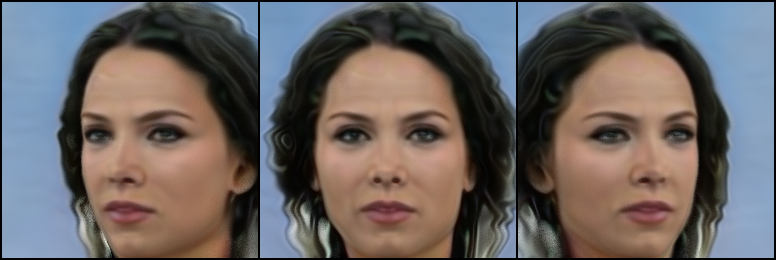} \\
&
\includegraphics[width=\linewidth]{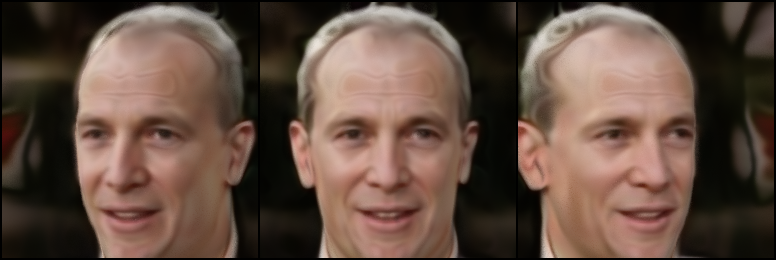} &
\includegraphics[width=\linewidth]{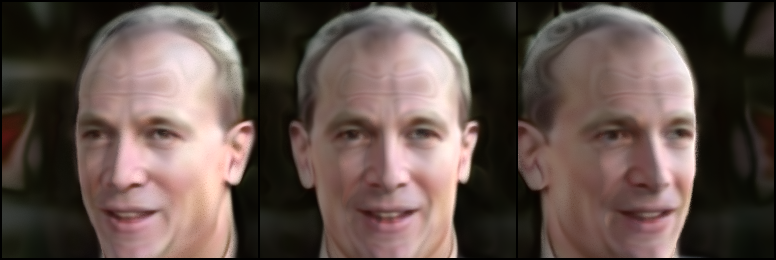} &
\includegraphics[width=\linewidth]{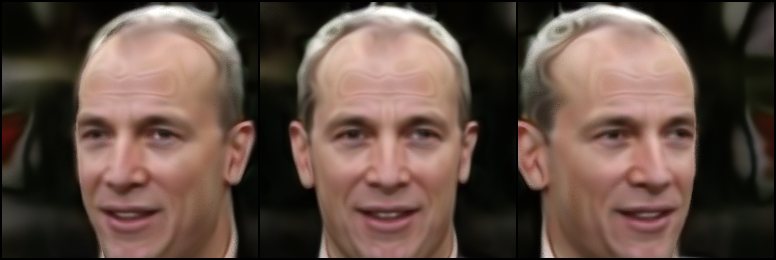} \\
\addlinespace
\multirow{2}{*}{\rotatebox[origin=c]{90}{VoluxGAN}}&
\includegraphics[width=\linewidth]{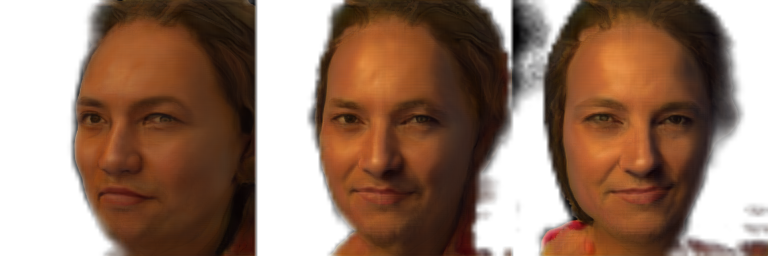} &
\includegraphics[width=\linewidth]{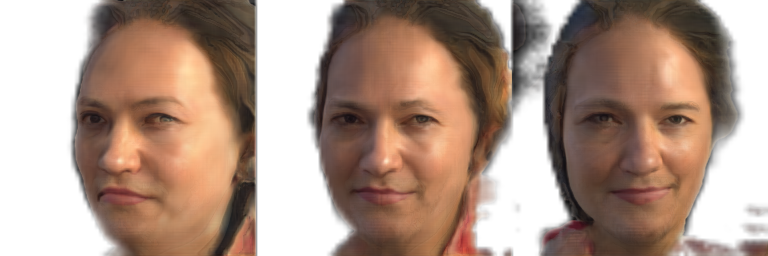} &
\includegraphics[width=\linewidth]{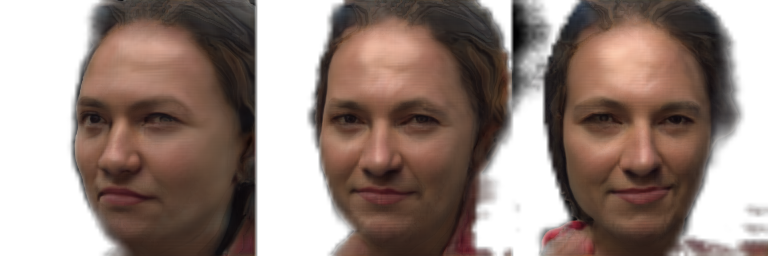} \\
&
\includegraphics[width=\linewidth]{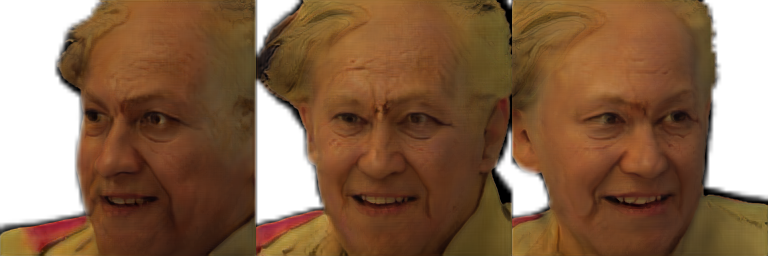} &
\includegraphics[width=\linewidth]{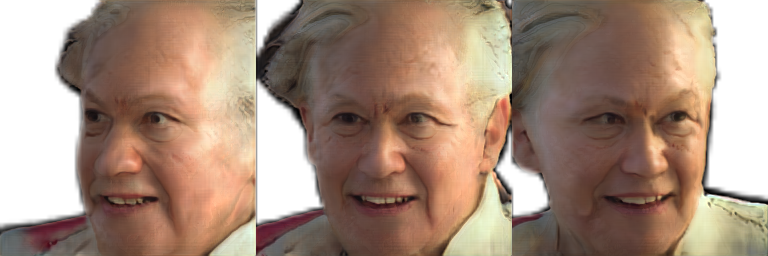} &
\includegraphics[width=\linewidth]{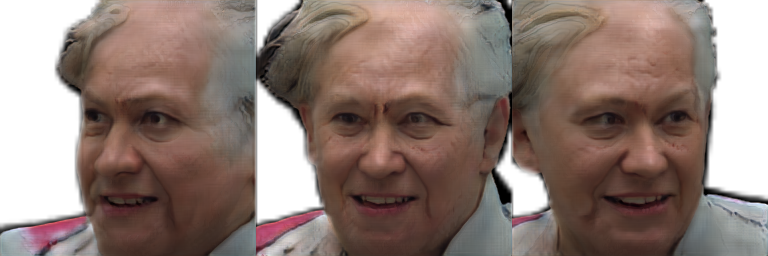} \\
\addlinespace
\multirow{2}{*}{\rotatebox[origin=c]{90}{\moniker{} (ours)}} &
\includegraphics[width=\linewidth]{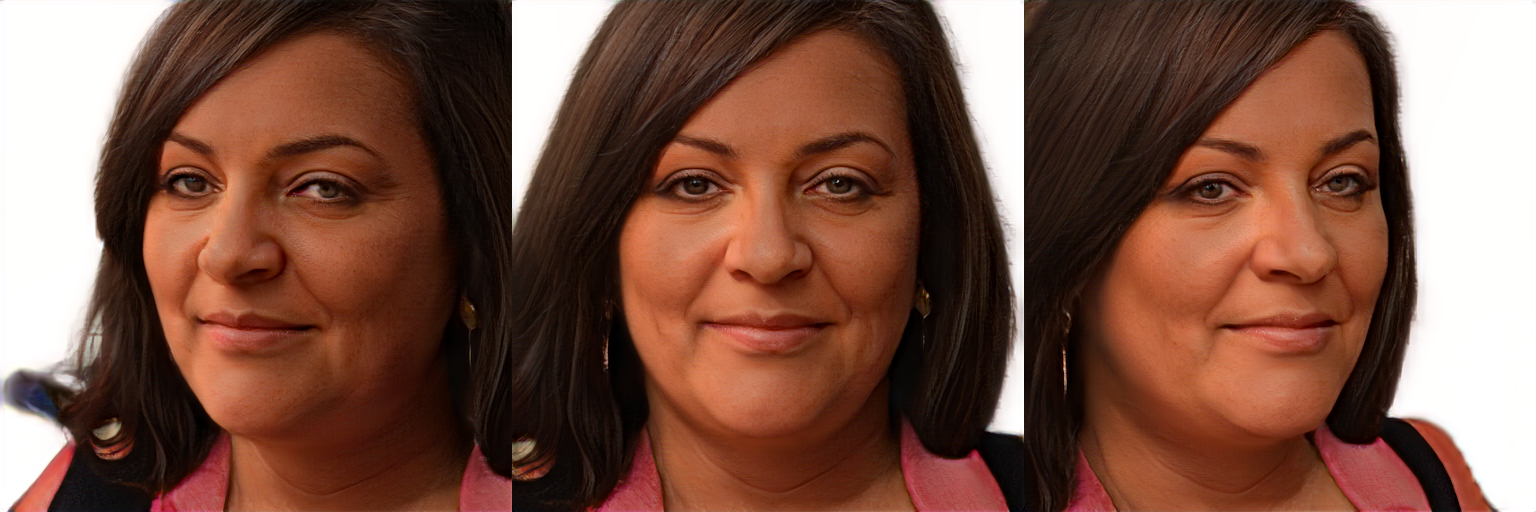} &
\includegraphics[width=\linewidth]{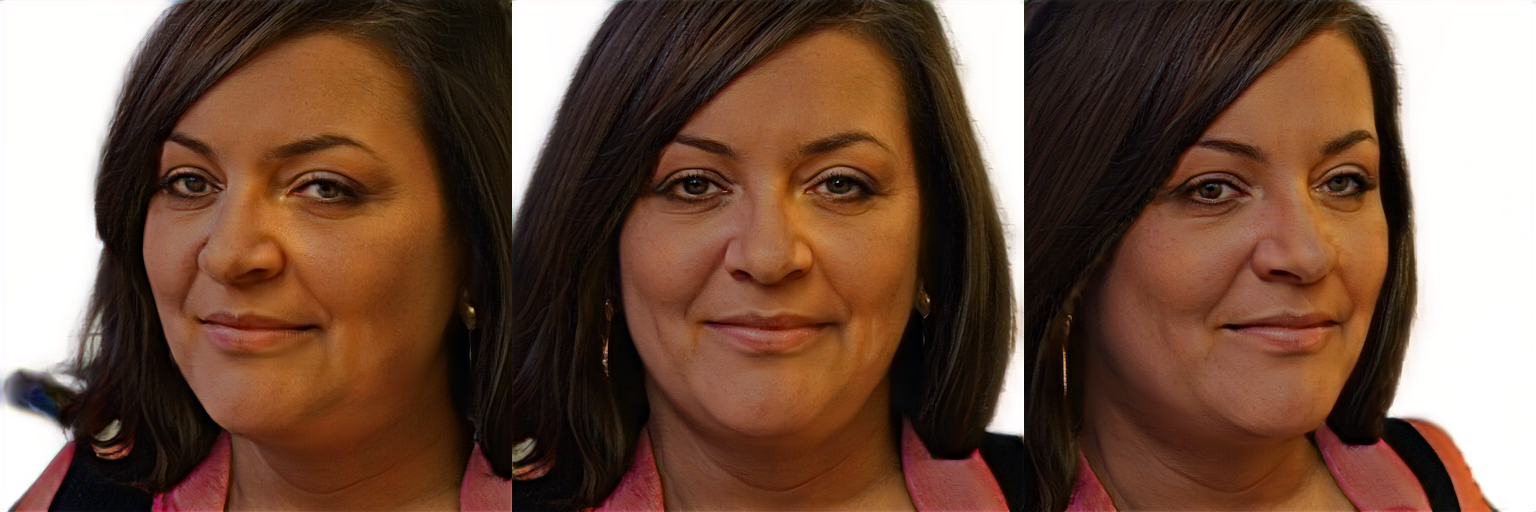} &
\includegraphics[width=\linewidth]{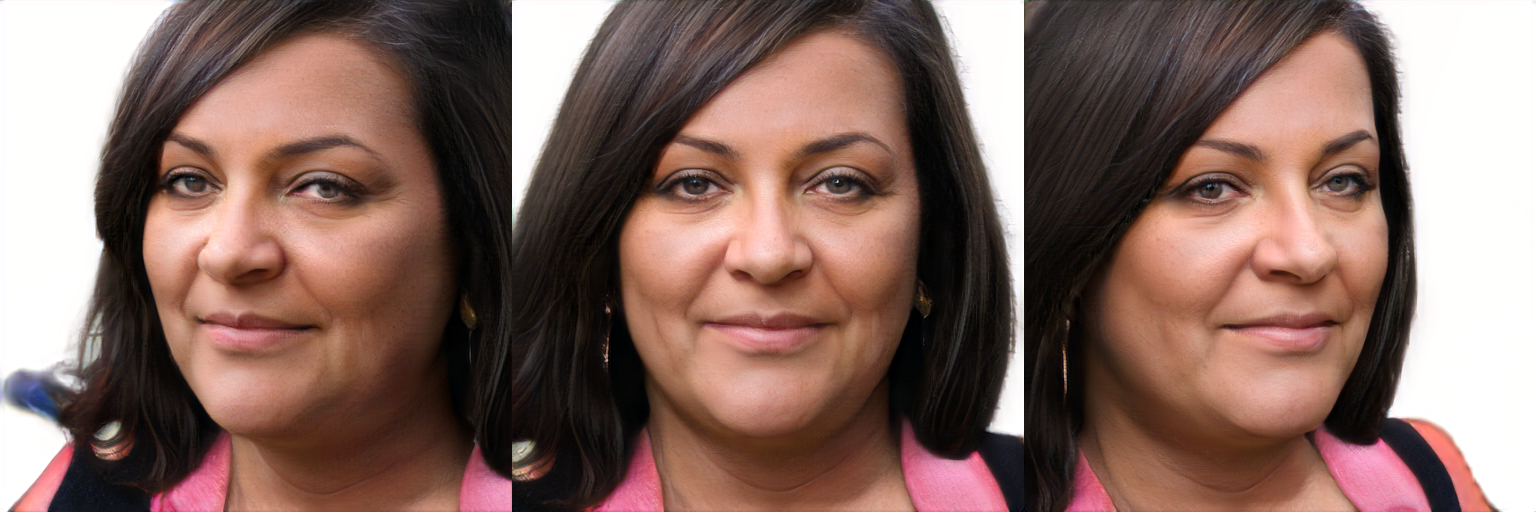} \\
&
\includegraphics[width=\linewidth]{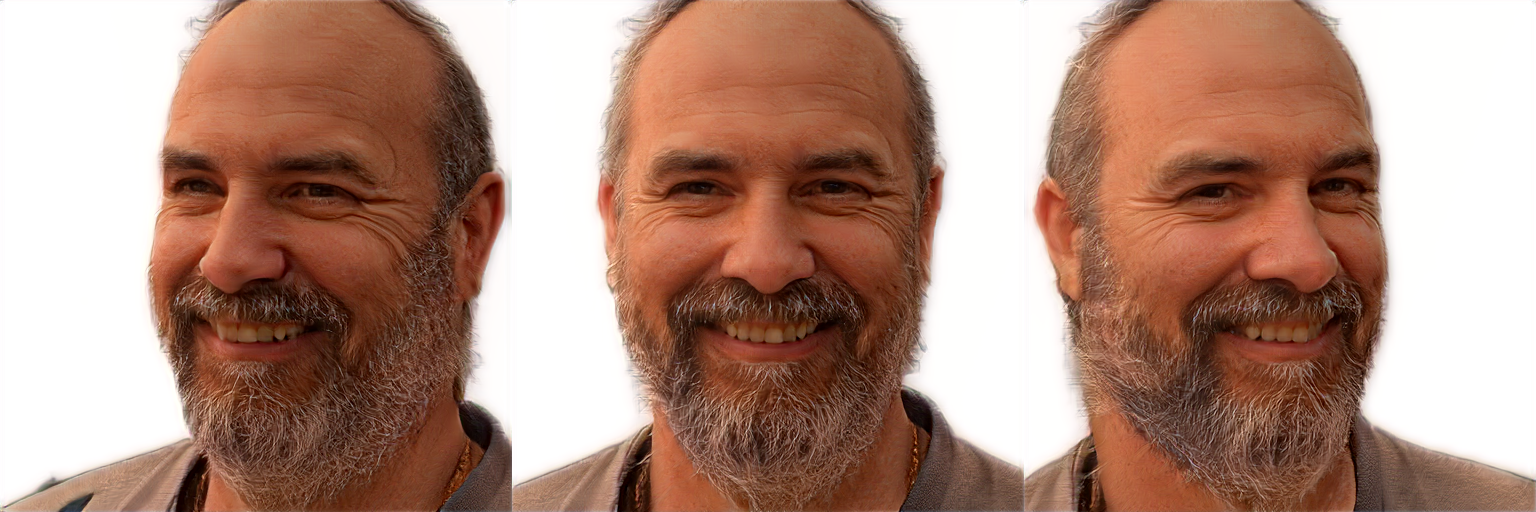} &
\includegraphics[width=\linewidth]{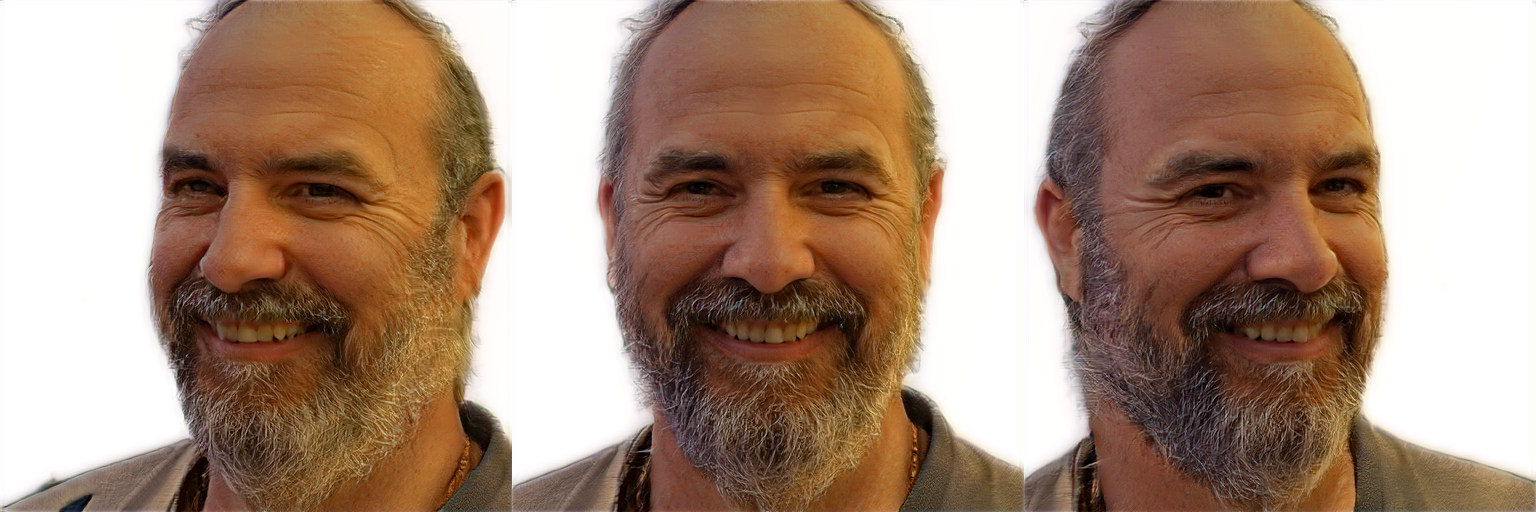} &
\includegraphics[width=\linewidth]{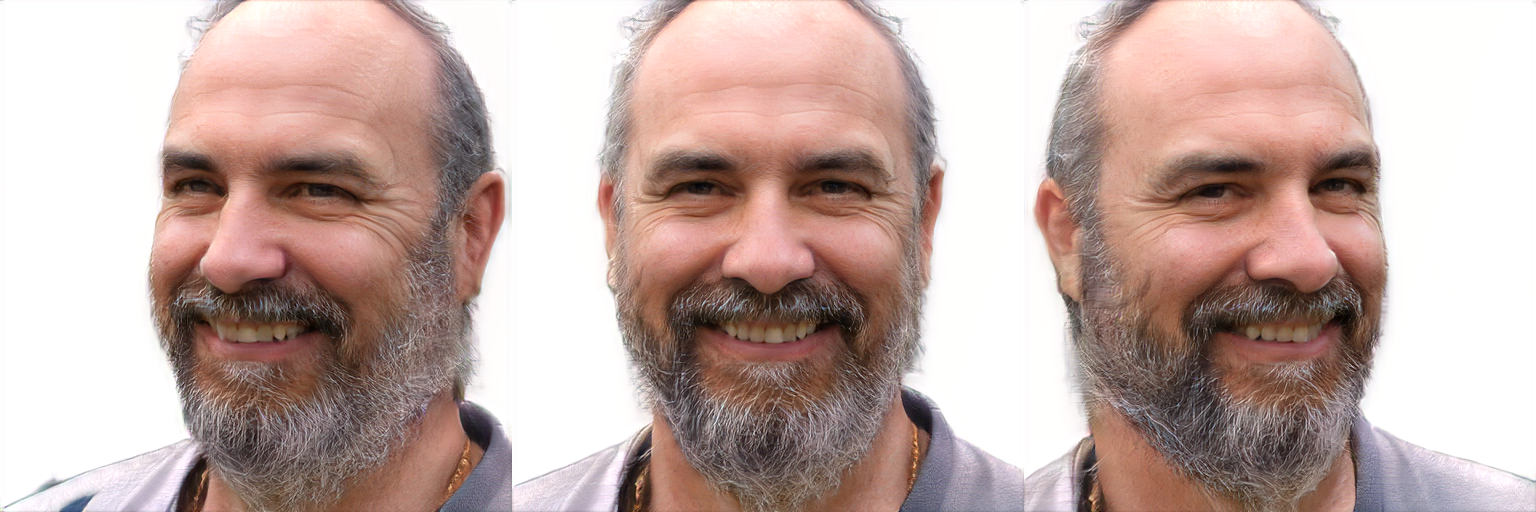} \\
\addlinespace
\raisebox{-.3\normalbaselineskip}[0pt][0pt]{\rotatebox[origin=c]{90}{Lighting}}
&
\includegraphics[width=\linewidth,height=0.33\linewidth]{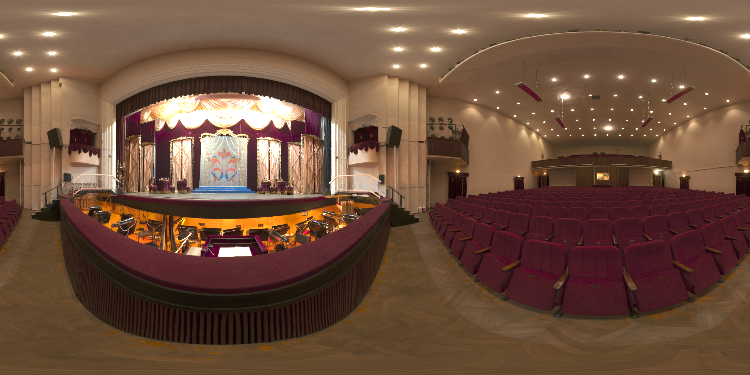} &
\includegraphics[width=\linewidth,height=0.33\linewidth]{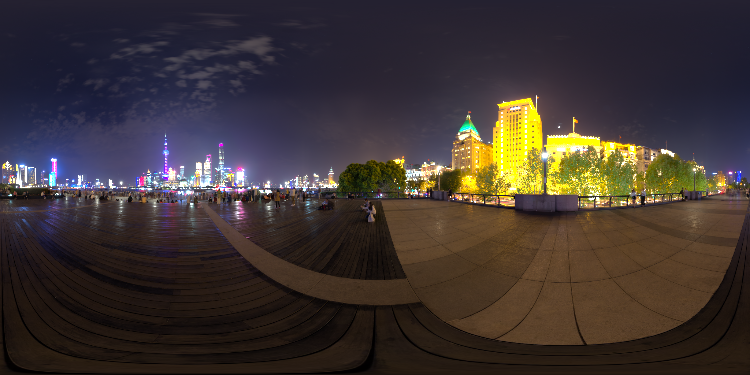} &
\includegraphics[width=\linewidth,height=0.33\linewidth]{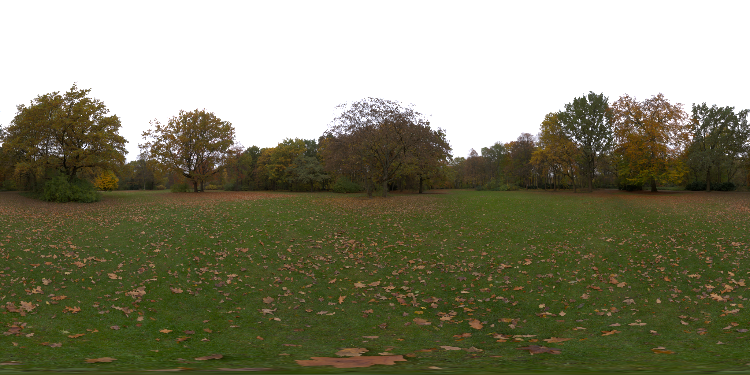} \\
\end{tabular}
\caption{\textbf{Qualitative Comparison.} We compare our generated 3D faces lit in various conditions with 2 baselines:  ShadeGAN~\cite{pan2021shading} and VoluxGAN~\cite{tan2022voluxgan}.
ShadeGAN uses a Lambertian lighting model and is limited to simple directional lights with changing hues, while VoluxGAN and \moniker{} use the environment maps shown in the last row.
Limited by its over-simplified lighting model, ShadeGAN cannot relight with complex lighting conditions.
Meanwhile, VoluxGAN's formulation ignores visibility and self-occlusion, which leads to disturbing artifacts, physically inaccurate reflections, and multi-view inconsistency.
Our \moniker{} generates more realistic multiview-consistent faces with plausible shadows under complex lighting conditions.}
\label{fig:gan_relight_compare}
\end{figure*}

\subsection{Comparison with state-of-the-art 3D GANs}\label{sec:sota}
\paragraph{Baselines.}
\moniker{} aims to achieve relightability while maintaining the photorealism demonstrated by existing non-relightable 3D GANs.
To this end, we compare \moniker{} with the state-of-the-art relightable 3D face generator, VoluxGAN~\cite{tan2022voluxgan}, as well as the state-of-the-art non-relightable 3D GANs, EG3D~\cite{chan2022efficient} and EpiGRAF~\cite{skorokhodov2022epigraf}, that represent the highest level of photorealism.
To maintain the same inference setup, we finetune EG3D with the proposed foreground mask discriminator at \(64\times64\) raw resolution to remove the background; this model is denoted as EG3D-noBG.

\noindent{\bf Visual Quality.}
To compare the visual quality of the lit images, we use Frechnet Inception Distance (FID)~\cite{heusel2017gans} and Kernel Inception Distance (KID)~\cite{binkowski2018demystifying}
following the protocol in ~\cite{chan2022efficient}.
Except for EpiGRAF and EG3D, the real images are filtered with the matting network~\cite{modnet2022ke} to remove the background.
As shown in~\cref{tbl:FID}, \moniker{} can achieve state-of-the-art FID and KID only marginally worse than EG3D and EG3D-noBG yet much better than the relightable alternative VoluxGAN.
The difference between our lighting distribution and the unknown actual FFHQ lighting distribution is a probable cause of the slight regression from EG3D.

\begin{figure}[htbp]
\centering \setlength{\tabcolsep}{1pt}
\renewcommand{\arraystretch}{0}%
\begin{tabular}{C{0.03\linewidth}C{0.31\linewidth}C{0.31\linewidth}C{0.31\linewidth}}
\rotatebox[origin=l]{90}{EG3D-noBG}&
\includegraphics[width=\linewidth]{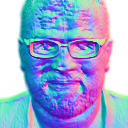} &
\includegraphics[width=\linewidth]{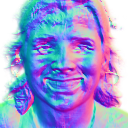} &
\includegraphics[width=\linewidth]{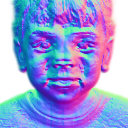} \\
\rotatebox[origin=l]{90}{VoluxGAN}&
\includegraphics[width=\linewidth]{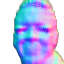} &
\includegraphics[width=\linewidth]{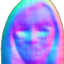} &
\includegraphics[width=\linewidth]{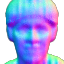} \\
\rotatebox[origin=l]{90}{\moniker{} (ours)}&
\includegraphics[width=\linewidth]{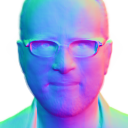} &
\includegraphics[width=\linewidth]{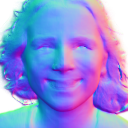} &
\includegraphics[width=\linewidth]{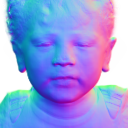} \\
\end{tabular}
\caption{\textbf{Geometry Quality.} \moniker's physically based lighting module depends on the normals, providing a strong yet unsupervised connection between geometry and generated RGB image.
This leads to less noisy geometry compared to EG3D. VoluxGAN~\cite{tan2022voluxgan} uses preconvolved light maps to enable relighting. This is less physically accurate than our approach for it omits visibility. Despite more aggressive supervision directly on the predicted normal maps, VoluxGAN is unable to generate as many details as ours.}
\label{fig:gan_shape_compare}
\end{figure}
\noindent{\bf Geometry Quality.}
The geometry of the generated faces is evaluated by comparing the output normal maps with references predicted from a pretrained facial normal predictor~\cite{Abrevaya_2020_CVPR}.
We compute the normal maps' cosine similarity as well as the L1 distance of their Laplacians to measure the fidelity of the geometric details~\cite{grassal2022neural}.
We also compute the identity similarity~\cite{deng2019arcface} between the frontal view and side views of varying yaw angles under fixed lighting to evaluate the view consistency.

As shown in \cref{tbl:FID}, \moniker{} yields more accurate normals than non-relightable approaches thanks to the stronger geometric cues induced by the geometry-dependent physically based lighting module (also see \cref{sec:ablation}).
An exception is VoluxGAN in the L1 Laplacian metric, which is likely biased for smooth inputs.
Indeed, the qualitative comparison in~\cref{fig:gan_shape_compare} shows that the normals of \moniker{} are more accurate and plausible, while the normals of VoluxGAN are overly smooth and the EG3D baseline contains many high-frequency artifacts. 
The identity similarity under varying viewing angles also reflects the geometry correctness.
\moniker{} compares favorably among the baselines, outperforming VoluxGAN by a large margin.
\begin{figure}[t!]
    \centering
\begin{subfigure}[t]{0.5\linewidth}
  \centering
  \includegraphics[width=0.96\linewidth]{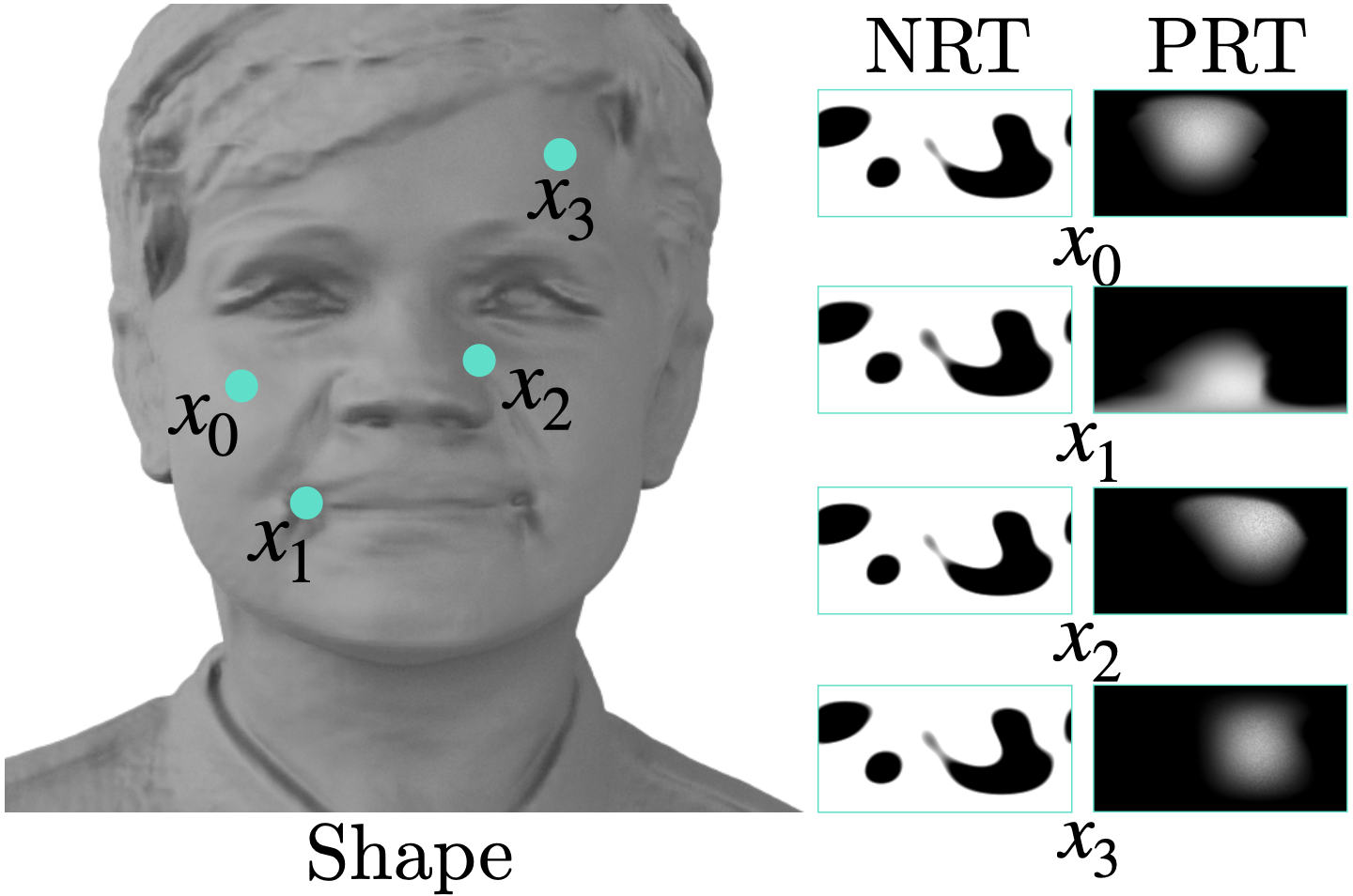}
  \caption{\textbf{Without} Self-Supervision.}
  \label{fig:without_selfsup}
\end{subfigure}%
\begin{subfigure}[t]{0.5\linewidth}
  \centering
  \includegraphics[width=0.96\linewidth]{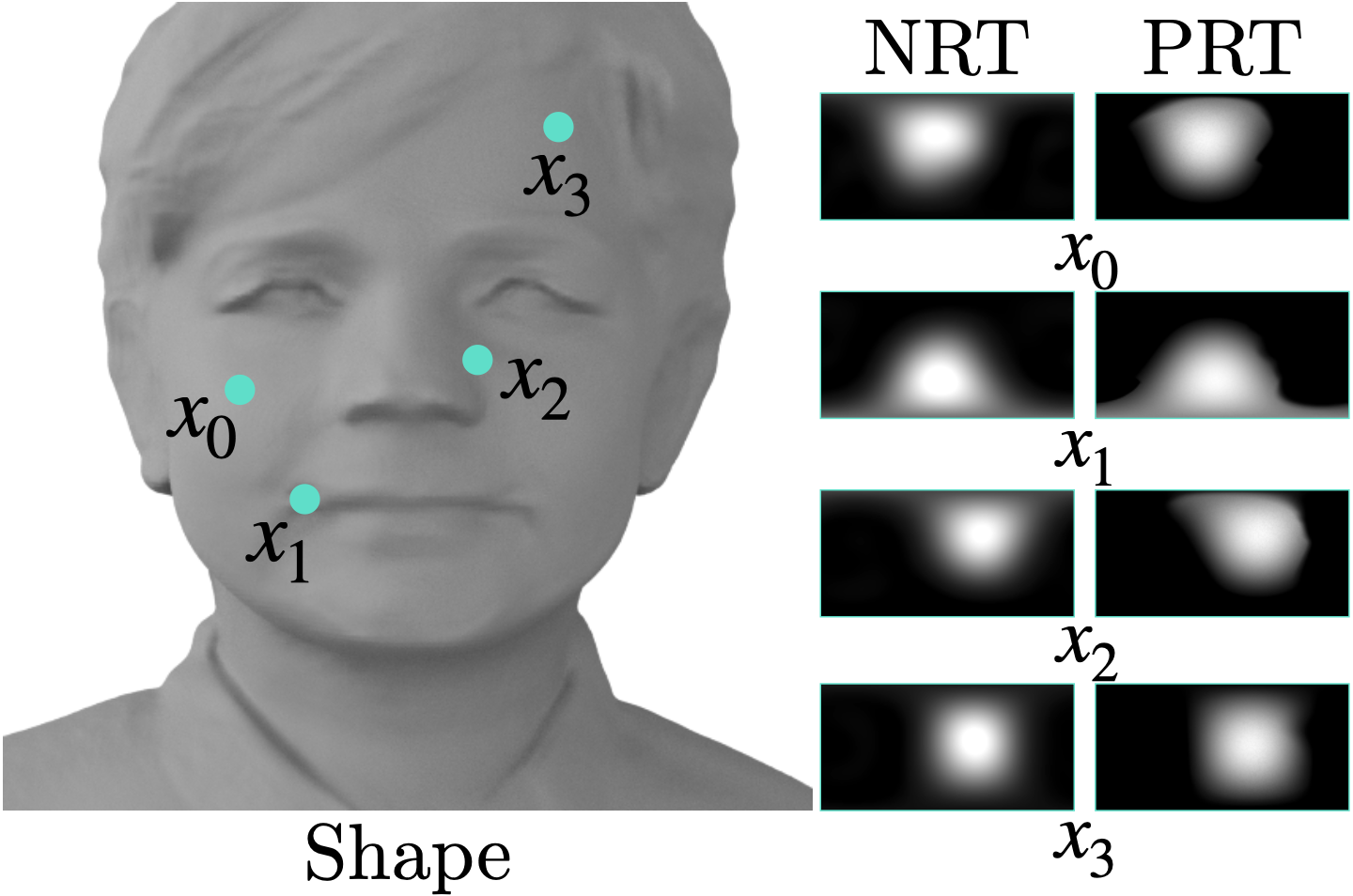}
  \caption{\textbf{With} Self-Supervision.}
  \label{fig:with_selfsup}
\end{subfigure}
\begin{subfigure}[t]{\linewidth}
  \centering
\resizebox{.96\linewidth}{!}{
\begin{tabular}{lccccc}
  \toprule
Location & $x_0$ & $x_1$ & $x_2$ & $x_3$ & Mean \\
\midrule
\textbf{Without} Self-Supervision   & $0.573$ & $0.568$ & $0.646$ & $0.672$ & $0.615$ \\
\textbf{With} Self-Supervision   & $\boldsymbol{0.008}$ & $\boldsymbol{0.004}$ & $\boldsymbol{0.008}$ & $\boldsymbol{0.002}$ & $\boldsymbol{0.005}$ \\
\bottomrule
\end{tabular}
} 
  \caption{Average L-2 error of predicted visibility map (NRT) and reference visibility map (PRT).}
  \label{fig:w_wo_sup_quant}
\end{subfigure}
    \caption{\textbf{Ablation.} We compare the learned NRT and the reference PRT with and without self-supervision by evaluating the visibility maps at several sampled surface points. With the proposed self-supervision (\cref{sec:autoprt}), the learned NRT is consistent with the reference PRT, thus resulting in geometrically consistent and physically accurate visibility prediction.
    Meanwhile, with this additional supervision signal from lighting, the generated geometry contains considerably fewer artifacts.}
    \label{fig:ablation}
\end{figure}

\noindent{\bf Relighting Quality.}
We use identity similarity between two images of the same person lit using two randomly sampled environment lightings to evaluate the relighting quality quantitatively.
The front-facing view is used for rendering and the score is averaged over 100 randomly generated people.
As \cref{tbl:FID} shows, \moniker{} significantly outperforms VoluxGAN in this metric.
This is demonstrated qualitatively in \cref{fig:gan_relight_compare}, where we also include ShadeGAN~\cite{pan2021shading} that uses a Lambertian lighting model, therefore is only relightable under simple directional light sources.
The relit images generated by \moniker{} demonstrate significantly higher visual quality.
VoluxGAN relies on the 2D convolution layers to add detail and shadows, which leads to strong inconsistencies between the relit images.
More results are available in the video and supplementary material.

\subsection{Ablation}\label{sec:ablation}
The self-supervision, \(\loss{nrt}\), enables efficient and geometrically consistent generation of shadows (\cref{sec:autoprt}).
The effectiveness of this term is demonstrated in \cref{fig:ablation}.
We visualize the geometry extracted from the generated density field, and compare the reference PRT and the predicted NRT at several points.
Specifically, the reference PRT is computed by ray tracing and projecting the resulting normal weighted visibility map onto the SH bases.
With our self-supervision, the predicted NRT closely matches the reference, i.e., the NRT is geometry consistent and physically interpretable.
Furthermore, we observe that the geometry generated with self-supervision contains fewer artifacts, indicating that imposing consistency between density and visibility leads to stronger supervision on geometry and better geometry -- appearance disambiguation, which significantly reduces artifacts seen from non-relightable 3D GANs.

\section{Discussion}\label{sec:discussion}

In summary, we propose an unconditional generative model, \moniker{}, to synthesize relightable 3D human faces, which not only yields lit faces with state-of-the-art photorealism but also considerably improves the geometry.
This is achieved using a physically based lighting module, which uses predicted geometry, albedo, specular tint, and visibility to render the diffuse and specular components of the face.
A key technical contribution is a novel approach to efficiently model visibility as neural radiance transfer during GAN training, which produces plausible shadows.
Together with a self-supervised training scheme, NRT directly predicts geometry-consistent visibility, thereby lifting secondary ray tracing and Monte Carlo integration otherwise needed to render realistic shadows.
Our generated results clearly excel in both photorealism and geometric quality compared to prior 3D GANs.

\paragraph{Limitations and Future Work.}
Similar to the original PRT, extending our NRT to dynamic scenes is non-trivial.
Combining relightability and animatability to achieve fully editable 3D human assets is an important future direction we would like to endeavor in the future.

\paragraph{Ethical Considerations.}
Our method could be misused to fabricate completely fictitious imagery or forge fake portraits of a real person, which can poses a societal threat.
We fervently condemn using our work with the intent of spreading misinformation or tarnishing reputation.
We also recognize a potential lack of diversity in our generated faces, stemming from implicit biases of the datasets we process.

\paragraph{Conclusion.}
\moniker{} is the first unconditional generative model that can achieve relightability while preserving the level of photorealism enabled by recent non-relightable approaches.
It is a vital step towards generating fully editable 3D human assets, allowing truly immersive experience in VR, AR, and film productions.

\paragraph{Acknowledgements.}
We thank Eric R. Chan for constructive discussions on training EG3D. We thank Alex W. Bergman, Manu Gopakumar, Suyeon Choi, Brian Chao, Ryan Po, and other members from the Stanford Computational Imaging Lab for helping on the computational resources and infrastructures. This project was in part supported by Samsung and Stanford HAI. Boyang Deng is supported by a Meta PhD Research Fellowship. Yifan Wang is supported by a Swiss Postdoc.Mobility Fellowship.

{\small
\bibliographystyle{ieee_fullname}
\bibliography{references}
}

\clearpage
\title{LumiGAN: Unconditional Generation of Relightable 3D Human Faces \\[.1in] Supplementary Material
}
\author{}
\date{}
\maketitle

\appendix

\setcounter{equation}{0}
\setcounter{figure}{0}
\setcounter{table}{0}
\makeatletter
\renewcommand{\theequation}{S\arabic{equation}}
\renewcommand{\thefigure}{S\arabic{figure}}

\section{Implementation Details}\label{sup:details}
This section describes the implementation details of our network architecture, training, and evaluation.

\subsection{Network Architecture}\label{supp:architecture}
\paragraph{Generator}
Our generator is based on the triplane generator proposed in  EG3D~\cite{chan2022efficient} and it follows the exact same architecture up to the triplane feature extraction.
Given the triplane feature \(\feature\left( \pnt \right)\) of a sample point \(\pnt\), we first use a 2-layer MLP decoder to predict the density \(\sigma\left( \pnt \right)\in\real\) and an intermediate 32-dimensional feature vector \(\feature_{\textrm{m}}\left( \pnt \right)\).
From the density, we compute the normal vector \(\normal\left( \pnt \right)\in\real^{3}\) (see Eq.9 in the main paper).
The normal vector and the aforementioned intermediate features are then concatenated and fed into another 2-layer MLP decoder to predict radiance transfer coefficients \(\left[ \prt_{1}\left( \pnt \right), \prt_{2}\left( \pnt \right), \dots, \prt_{\shdeg}\left( \pnt \right) \right]\), where \(\shdeg\) is set to 25, corresponding to the first four degrees of SH bases.
In parallel, a fully-connected layer with \textsc{softplus} activation transforms the intermediate feature vector \(\feature_{\textrm{m}}\left( \pnt \right)\) to the diffuse albedo \(\brdf^{d}\left( \pnt \right)\in\real^{3}\)and specular tint  \(\spectint\left( \pnt \right)\in\real\).
These quantities are then used in our lighting module (see Eq.6 and 7 in the main paper) to obtain the diffuse and specular radiances, \(\outrdnc^{d}\left( \pnt \right)\) and \(\outrdnc^{s}\left( \pnt,\outview \right)\), from which the final radiance \(\outrdnc\left( \pnt,\outview \right)\) is obtained as the sum of the two.
The volume rendering equation (Eq. 8) is applied on this final radiance \(\outrdnc\left( \pnt,\outview \right)\) and the intermediate feature \(\feature_{\textrm{m}}\) to obtain a low-resolution raw image and feature map respectively.
Finally, the raw image is converted from the linear space to the sRGB space~\cite{wikisrgb} and, together with the feature map, fed to the EG3D super-resolution module, which first bilinearly upsamples inputs to \(128\times128\) and then applies several neural filters to obtain final \(512\times512\) high-resolution RGB image.

\paragraph{Discriminators}
As described Sec. 3.3 in the main paper, we use an additional discriminator to encourage generating foreground-only portraits.
The discriminator follows the design of the EG3D discriminator except that it takes the low-resolution accumulated weights from the volume renderer as fake input and the preprocessed foreground masks of the dataset as real input, whereas the image discriminator is the same as the one used in EG3D~\cite{chan2022efficient}.

\subsection{Training}\label{supp:training}
\paragraph{Data}
As mentioned in the main paper, we use real-world environment lighting to train our model.
For this purpose, we collected 459 HDR environment maps from Poly Haven~\cite{polyheaven} for training.
Each map is projected onto the first 4 degrees of SH bases, resulting in 25 SH coefficients, which we use as input to our lighting module.

\paragraph{Losses}
We define our training objectives based on EG3D.
Besides the image GAN loss~\cite{goodfellow2014generative} and the R1 density regularization~\cite{mescheder2018training} used in EG3D, our training loss contains additionally the mask GAN loss (defined the same as the image GAN loss), the NRT consistency loss \(\loss{nrt}\) (Eq. 11) and the albedo smoothness regularization \(\loss{smooth}\) (Eq. 12).
The total loss is the sum of the aforementioned losses, with the weight of the NRT consistency loss set to \(50.0\) and the rest to \(1.0\).

In particular, for the NRT consistency loss, we sample a total of 10 rays, including the two primary rays \(\outview\) and \(-\outview\), and 8 auxiliary rays randomly sampled in the negative hemisphere, \ie, \(\normal\left( \pnt \right)\cdot\inview < 0\).
Note, as we explained in the paper, the reference visibility of the primary rays involves minimal additional computational cost, as the sample densities required to compute the visibility are already evaluated for volume rendering.
For the auxiliary rays in the negative hemisphere, since by definition \(\normatten\left( \pnt, \inview \right) \coloneqq \max(0, \normal( \pnt ) \cdot \inview) \equiv 0\), the reference PRT is always zero.

\paragraph{Training Strategy}
Instead of training our model from scratch, we initialize our generator with a pre-trained EG3D model and keep the same training hyper-parameters.
More specifically, we first warm up the training of our NRT prediction modules while freezing the weights of the rest of the pipeline for $10$k iterations, using only the EG3D losses and our NRT consistency loss $\loss{nrt}$.
Then we fine-tune all weights using the same losses for another $10$k iterations.
At last, we introduce the remaining loss terms $loss{smooth}$ and train the whole model for $25$k iterations.

\subsection{Evaluation}\label{supp:evaluation}
\paragraph{Baselines.}
The baselines for our quantitative evaluation in Tab. 1 of the main paper include state-of-the-art methods that are trained using FFHQ dataset.

\paragraph{Metrics.}
Here, we provide more details about the metrics used in our evaluation.

\noindent\emph{\underline{FID and KID.}}
When available, we show the FID and KID evaluations provided in the original publications, \eg EpiGRAF~\cite{skorokhodov2022epigraf} and EG3D~\cite{chan2022efficient}; otherwise, we compute the FID and KID evaluations based on the same settings as in EG3D~\cite{chan2022efficient}, in which the feature statistics of 50K generated images are compared with that of the entire training dataset (\(512\times512\) FFHQ images preprocessed as per EG3D).

When the method generates images in a resolution different from the FFHQ dataset, \eg VoluxGAN, the dataset is bicubicly downsampled to the matching resolution before feature extraction.
For our method and VoluxGAN~\cite{tan2022voluxgan}, which produce background-free portraits, we filter the background of the training dataset using an off-the-shelf matting network~\cite{modnet2022ke} to obtain comparable feature statistics.
In particular, since VoluxGAN by default generates images with black backgrounds, we also the composite black background for the training dataset when evaluating VoluxGAN.
Note, the same matting network is used to generate the foreground masks for our mask discriminator.

\noindent\emph{\underline{Normal Accuracy.}}
We evaluate the geometry quality by comparing generated normal maps computed from the density field and reference normal maps predicted by the off-the-shelf face normal network~\cite{Abrevaya_2020_CVPR}.
The evaluation includes 500 normal maps synthesized at randomly sampled viewing angles.
In particular, we follow the procedure in~\cite{grassal2022neural} to generate accurate reference normal maps: the generated RGB image is first cropped to the face bounding box detected by~\cite{bulat2017far}, then padded to square and resized to \(256\times256\) before feeding to the face normal network.
For the generated normal maps, they are first rotated to the  camera coordinates, then applied the same cropping, padding and resizing steps.

As described in the main paper, we use two metrics to evaluate the normal accuracy: the cosine similarity and L1 distance of the Laplacian of the normal maps, defined respectively below
\begin{gather}
\frac{1}{LM}\sum_{l,m}\left(\normalmap_{l,m} \cdot\normalmap^{\textrm{ref}}_{l,m}\right)\circ \mask_{l,m}\\
\frac{1}{LM}\sum_{l,m}\left|\left(\Lap\normalmap)_{l,m} - (\Lap\normalmap^{\textrm{ref}}\right)_{l,m}\right|\circ \mask_{l,m},
\end{gather}
where \(\normalmap_{l,m}\) and \(\normalmap^{\textrm{ref}}_{l,m}\) are the normal map and reference normal map at pixel index \(l, m\), respectively, \(\Lap\) is the Laplacian filter implemented as the difference between the input image and itself after Gaussian blur, \(\mask_{l,m}\) is face segmentation mask from~\cite{lin2022robust}.

\noindent\emph{\underline{Identity Similarity under view and lighting.}}
We evaluate the identity similarity score~\cite{deng2019arcface} under view and lighting changes to measure the coarse geometry quality and the relighting quality, respectively.
For changing the view angle, we randomly generate 100 identities under the same environment lighting but render with \(\left\{-0.5, 0.25, 0, 0.25, 0.5\right\}\) radian rotation around the vertical axis, and compare the rotated images with the front-facing image.
For changing lighting, we randomly synthesize 100 front-facing identities but render them under two randomly sampled environment maps, then we compare the identity similarity between the two differently lit images.

\section{Further Experiments and Results}\label{supp:experiments}
\subsection{Quantitative evaluation for ShadeGAN}\label{supp:full_fid}
\begin{table*}[htbp]
    \centering\begin{tabular}{ccc*{2}{c}*{4}{c}c}
\toprule
\multirow{2}{*}{method, resolution} & \multirow{2}{*}{FID $\downarrow$} & \multirow{2}{*}{KID $\downarrow$} & \multicolumn{2}{c}{Normal} & \multicolumn{4}{c}{Id. Similarity (view) $\uparrow$} & \multirowcell{2}{Id. Similarity\\(lighting) $\uparrow$}\\\cmidrule(lr){4-5}\cmidrule(lr){6-9}
& & & Cos $\uparrow$ & LapL1 $\downarrow$ & -0.5 & -0.25 & 0.25 & 0.5 \\\midrule
ShadeGAN, \(128^{2}\) & \(9.71^{\color{red} *}\) & \(4.880^{\color{red} *}\) & \textbf{0.86} & \textbf{0.031} & 0.665 & 0.861 & 0.837 & 0.648 & 0.822 \\
\midrule
VoluxGAN, \(256^{2}\) & 59.79 & 4.124 & 0.78 & \underline{0.033} & 0.606 & 0.774 & 0.800 & 0.599 &  0.831 \\
EpiGRAF, \(512^{2}\) & 9.92 & 0.453 & 0.75 & 0.098 & \underline{0.756} & \textbf{0.911} & \textbf{0.910} & \underline{0.735} & -   \\
EG3D, \(512^{2}\) & \underline{4.70} & \textbf{0.132} & 0.73 & 0.088 & 0.743 & 0.814 & 0.812 & 0.741 & - \\
EG3D-noBG, \(512^{2}\) & \textbf{4.59} & \underline{0.200} & 0.75 & 0.084 & 0.684 & 0.782 & 0.795 & 0.704 & -  \\
\moniker{} (ours), \(512^{2}\) & {5.28} & {0.251} & \underline{0.79} & 0.060 & \textbf{0.765} & \underline{0.888} &  \underline{0.883} & \textbf{0.772} & \textbf{0.947}
\\\bottomrule
\end{tabular}
    \caption{Quantitative evaluation including ShadeGAN. {\color{red} *}Note that the publicly available ShadeGAN model is trained on CelebA~\cite{liu2015faceattributes} and thus the FID and KID are evaluated using CelebA dataset, therefore these results are not directly comparable with those of other methods.}
    \label{tbl:shadegan_eval}
\end{table*}
We did not include ShadeGAN~\cite{pan2021shading} in Tab. 1 of our main paper because it is trained on CelebA~\cite{liu2015faceattributes} and thus the FID and KID scores are evaluated using CelebA dataset. Therefore, these results are not directly comparable with those of other methods.
Here, We include the results of ShadeGAN in \cref{tbl:shadegan_eval} for completeness.
At the same time, \cref{tbl:shadegan_eval} suggests that ShadeGAN and VoluxGAN produce the best normal qualitatively.
Yet from the visual comparison below, we can see that the normal maps produced by VoluxGAN contain obvious artifacts, and our normal maps have more geometric details.
\begin{figure}[htbp]
\centering \setlength{\tabcolsep}{1pt}
\renewcommand{\arraystretch}{0}%
\begin{tabular}{C{0.03\linewidth}C{0.31\linewidth}C{0.31\linewidth}C{0.31\linewidth}}
\rotatebox[origin=l]{90}{EG3D-noBG}&
\includegraphics[width=\linewidth]{figure/normal_result/eg3d/seed-0021_normal.png} &
\includegraphics[width=\linewidth]{figure/normal_result/eg3d/seed-0043_normal.png} &
\includegraphics[width=\linewidth]{figure/normal_result/eg3d/seed-0092_normal.png} \\
\rotatebox[origin=l]{90}{ShadeGAN}&
\includegraphics[width=\linewidth]{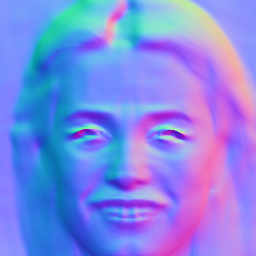} &
\includegraphics[width=\linewidth]{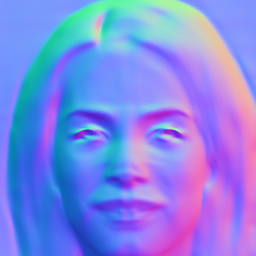} &
\includegraphics[width=\linewidth]{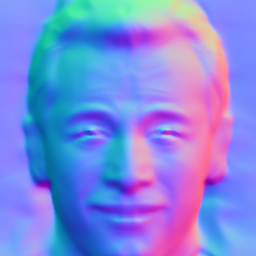} \\
\rotatebox[origin=l]{90}{VoluxGAN}&
\includegraphics[width=\linewidth]{figure/normal_result/voluxGAN/00010_normal.png} &
\includegraphics[width=\linewidth]{figure/normal_result/voluxGAN/00017_normal.png} &
\includegraphics[width=\linewidth]{figure/normal_result/voluxGAN/00049_normal.png} \\
\rotatebox[origin=l]{90}{\moniker{} (ours)}&
\includegraphics[width=\linewidth]{figure/normal_result/ours/seed-0021_normal.png} &
\includegraphics[width=\linewidth]{figure/normal_result/ours/seed-0043_normal.png} &
\includegraphics[width=\linewidth]{figure/normal_result/ours/seed-0092_normal.png} \\
\end{tabular}
\caption{\textbf{Normal comparison (extended).} All normal maps are rendered from the volumetric renderer at \(128\times128\) resolution. \moniker{} and ShadeGAN produce normals with significantly higher quality, while \moniker{} has visibly more geometric details.}
\label{fig:norma_comp_with_shadegan}
\end{figure}
\subsection{Effect of albedo smoothness regularization.}\label{supp:albedo_smoothness}
\begin{figure}[htbp]
    \centering
    \setlength{\tabcolsep}{0pt}
    \renewcommand{\arraystretch}{0}
    \begin{tabular}{c*{3}{C{0.33\linewidth}}}
    & albedo & specular & rendering \\
    \rotatebox[origin=c]{90}{without $\loss{smooth}$} &
    \includegraphics[width=\linewidth]{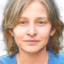} &
    \includegraphics[width=\linewidth]{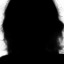} &
    \includegraphics[width=\linewidth]{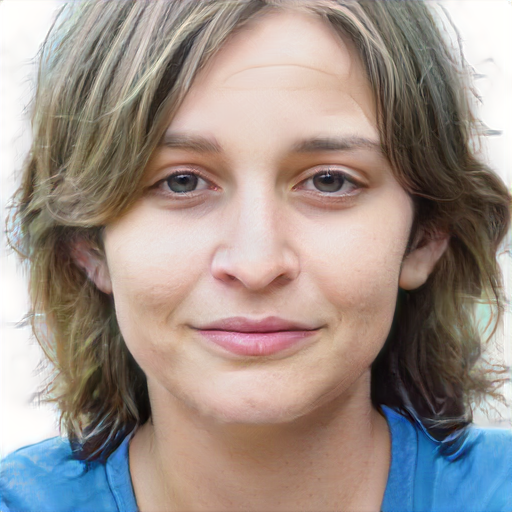} \\
    \rotatebox[origin=c]{90}{with $\loss{smooth}$} &
    \includegraphics[width=\linewidth]{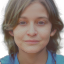} &
    \includegraphics[width=\linewidth]{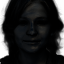} &
    \includegraphics[width=\linewidth]{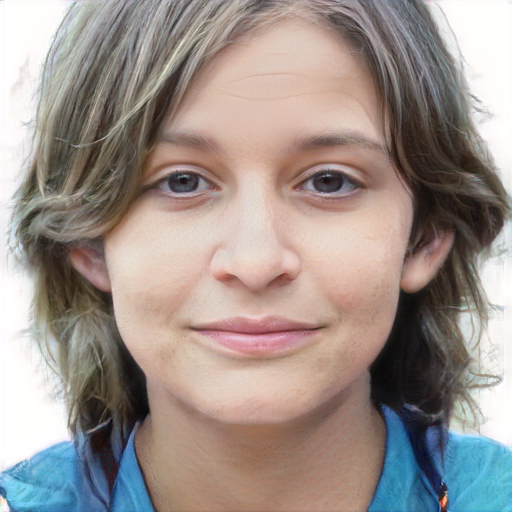} \\
    \end{tabular}
    \caption{Albedo smoothness regularization. Without the regularization, the specular highlights are leaked into the albedo, and the specular radiance is not well captured.}
    \label{fig:albedo_smoothness}
\end{figure}
\cref{fig:albedo_smoothness} shows the effect of albedo smoothness regularization.
Without the regularization, the specular highlights are leaked into the albedo, the specular radiance is not well captured.
\subsection{Uncurrated qualitative results.}
\begin{figure*}[htbp]
    \centering
    \renewcommand{\arraystretch}{0}%
    \setlength\tabcolsep{0pt}
    \newlength{\imgheight}
    \setlength{\imgheight}{0.132\linewidth}
    \begin{tabular}{C{0.06\linewidth}*{7}{C{\imgheight}}}
    \multicolumn{8}{c}{{ShadeGAN}~\cite{pan2021shading}:} \\
    &
    \includegraphics[width=\linewidth]{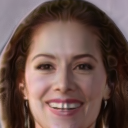} &
    \includegraphics[width=\linewidth]{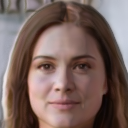} &
    \includegraphics[width=\linewidth]{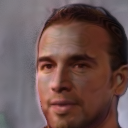} &
    \includegraphics[width=\linewidth]{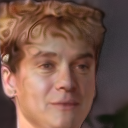} &
    \includegraphics[width=\linewidth]{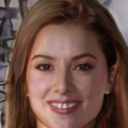} &
    \includegraphics[width=\linewidth]{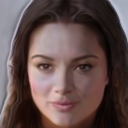} &
    \includegraphics[width=\linewidth]{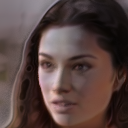} \\
    &
    \includegraphics[width=\linewidth]{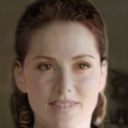} &
    \includegraphics[width=\linewidth]{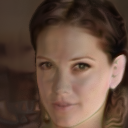} &
    \includegraphics[width=\linewidth]{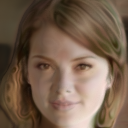} &
    \includegraphics[width=\linewidth]{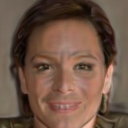} &
    \includegraphics[width=\linewidth]{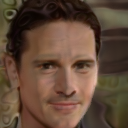} &
    \includegraphics[width=\linewidth]{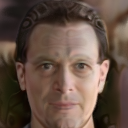} &
    \includegraphics[width=\linewidth]{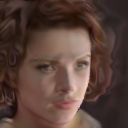} \\\addlinespace
    \multicolumn{8}{c}{{VoluxGAN}~\cite{tan2022voluxgan}:} \\
    \frame{\includegraphics[height=\linewidth,width=\imgheight,angle=90,origin=c]{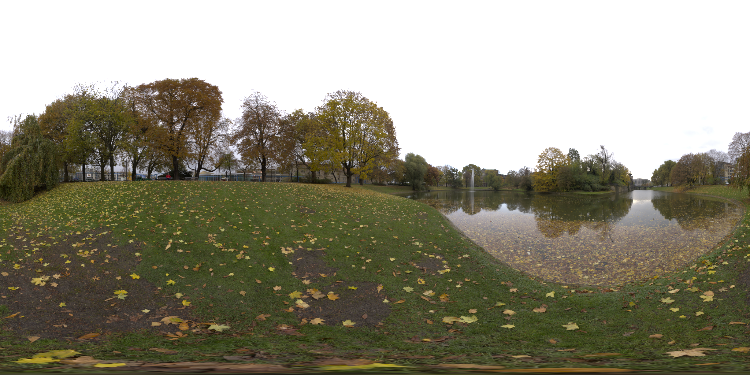}} &
    \includegraphics[width=\linewidth]{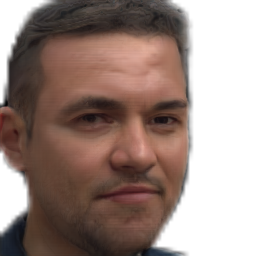} &
    \includegraphics[width=\linewidth]{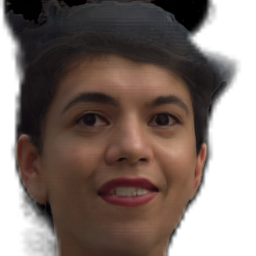} &
    \includegraphics[width=\linewidth]{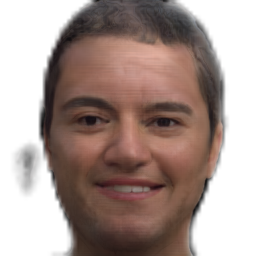} &
    \includegraphics[width=\linewidth]{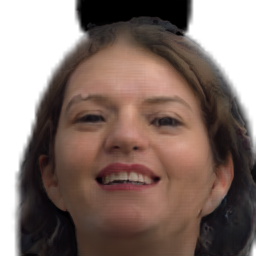} &
    \includegraphics[width=\linewidth]{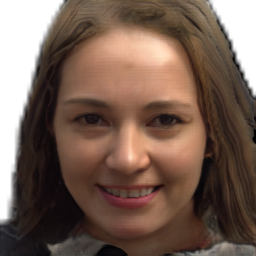} &
    \includegraphics[width=\linewidth]{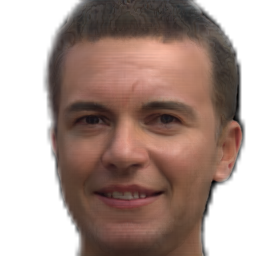} &
    \includegraphics[width=\linewidth]{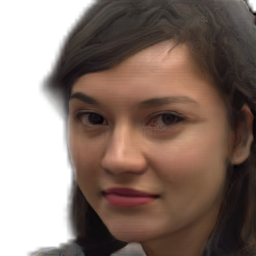}
    \\
    \frame{\includegraphics[height=\linewidth,width=\imgheight,angle=90,origin=c]{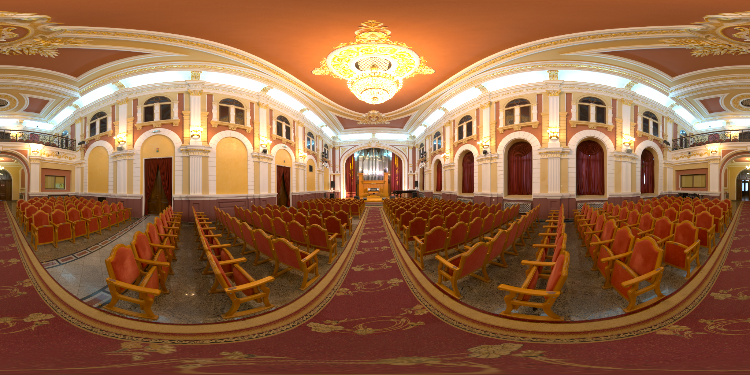}} &
    \includegraphics[width=\linewidth]{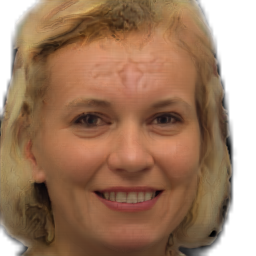} &
    \includegraphics[width=\linewidth]{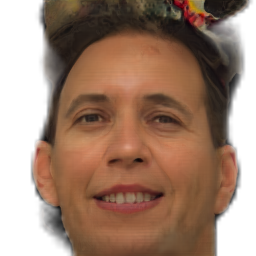} &
    \includegraphics[width=\linewidth]{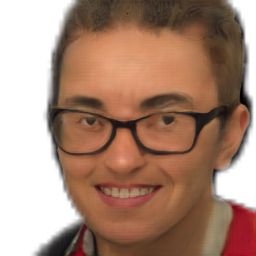} &
    \includegraphics[width=\linewidth]{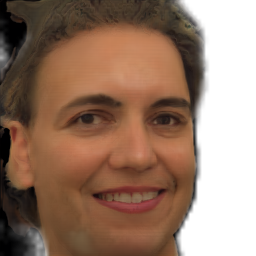} &
    \includegraphics[width=\linewidth]{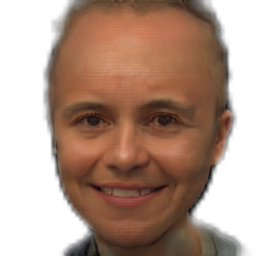} &
    \includegraphics[width=\linewidth]{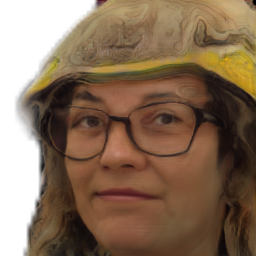} &
    \includegraphics[width=\linewidth]{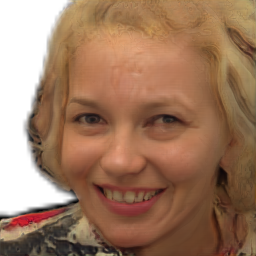}
    \\
    \frame{\includegraphics[height=\linewidth,width=\imgheight,angle=90,origin=c]{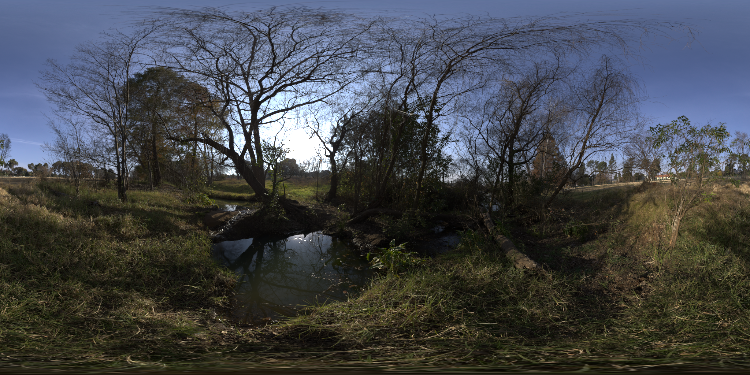}} &
    \includegraphics[width=\linewidth]{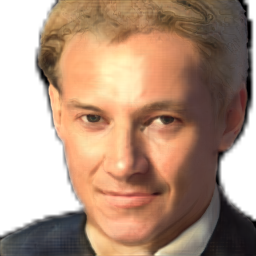} &
    \includegraphics[width=\linewidth]{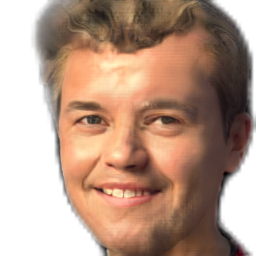} &
    \includegraphics[width=\linewidth]{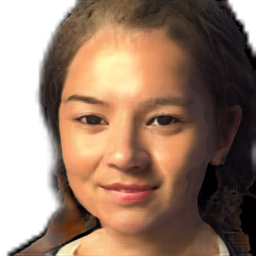} &
    \includegraphics[width=\linewidth]{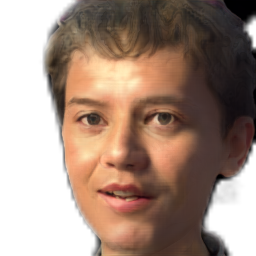} &
    \includegraphics[width=\linewidth]{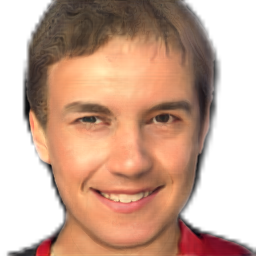} &
    \includegraphics[width=\linewidth]{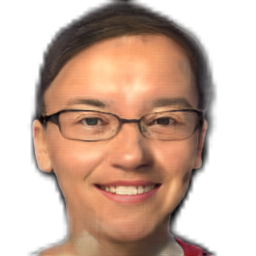} &
    \includegraphics[width=\linewidth]{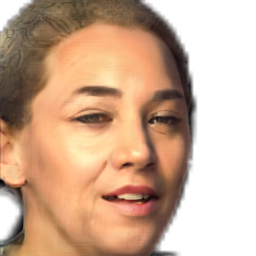}
    \\\addlinespace
    \multicolumn{8}{c}{\moniker{} (ours):} \\
    \frame{\includegraphics[height=\linewidth,width=\imgheight,angle=90,origin=c]{figure/paper_uncurated/hdri_library/dresden_moat_2k_raw.png}} &
    \includegraphics[width=\linewidth]{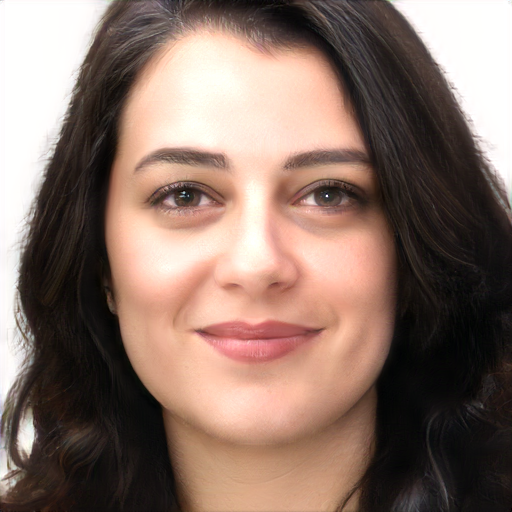} &
    \includegraphics[width=\linewidth]{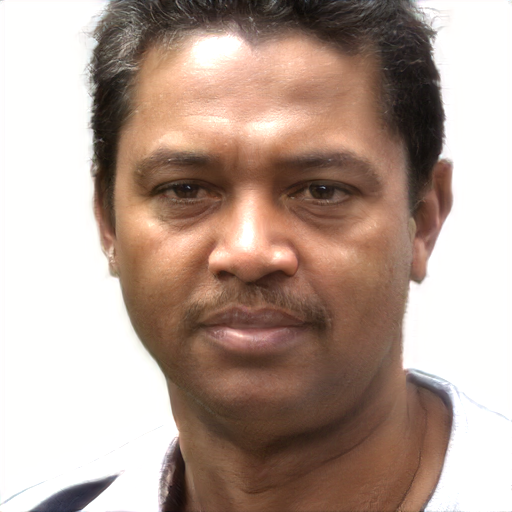} &
    \includegraphics[width=\linewidth]{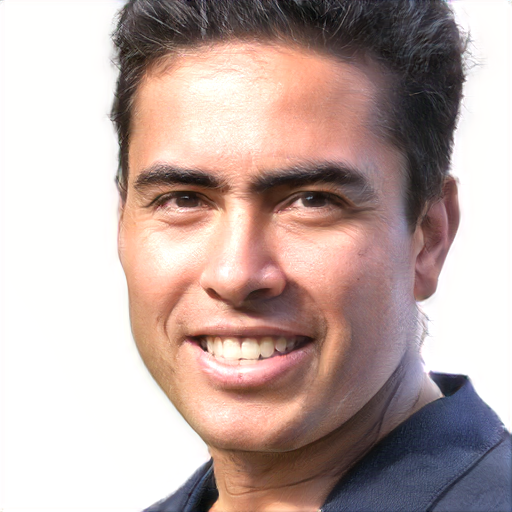} &
    \includegraphics[width=\linewidth]{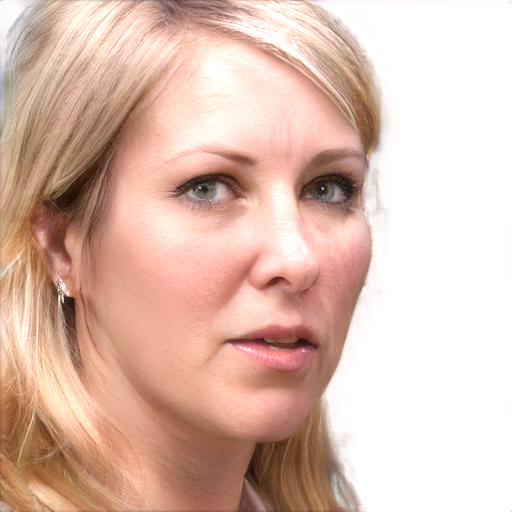} &
    \includegraphics[width=\linewidth]{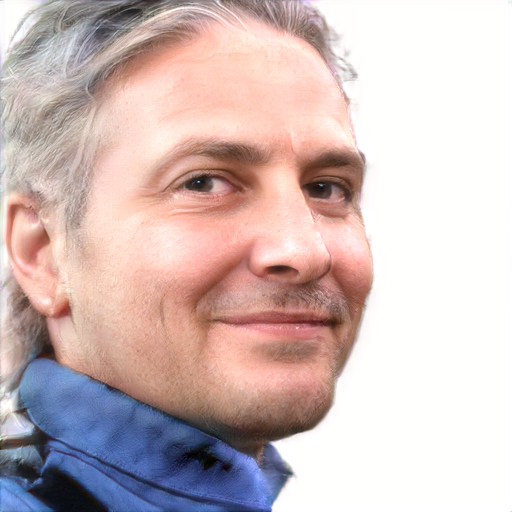} &
    \includegraphics[width=\linewidth]{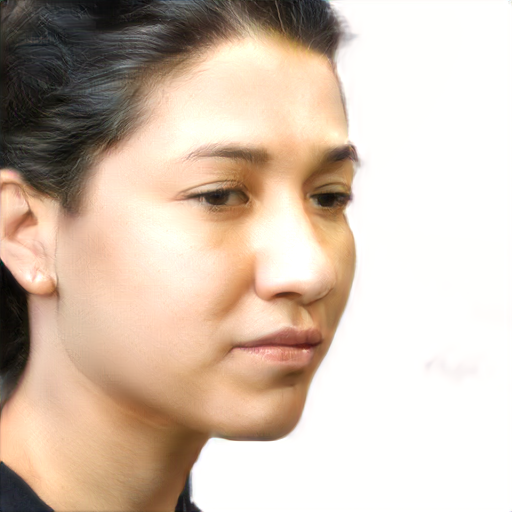} &
    \includegraphics[width=\linewidth]{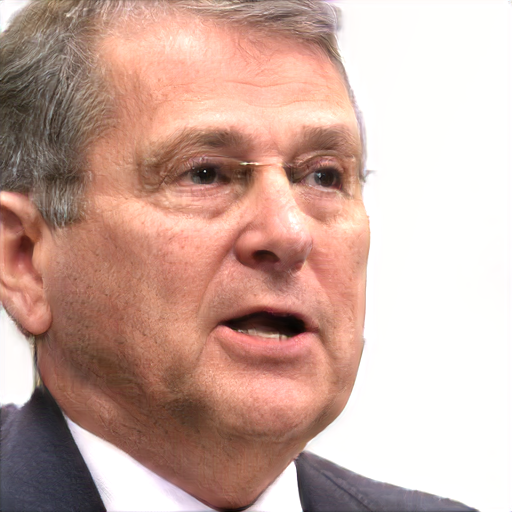}
    \\
    \frame{\includegraphics[height=\linewidth,width=\imgheight,angle=90,origin=c]{figure/paper_uncurated/hdri_library/music_hall_01_2k_raw.png}} &
    \includegraphics[width=\linewidth]{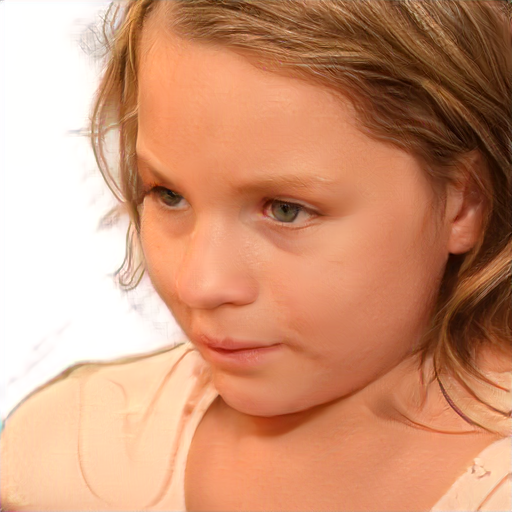} &
    \includegraphics[width=\linewidth]{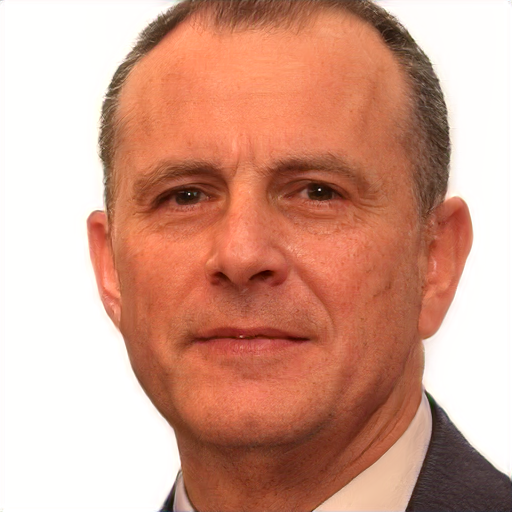} &
    \includegraphics[width=\linewidth]{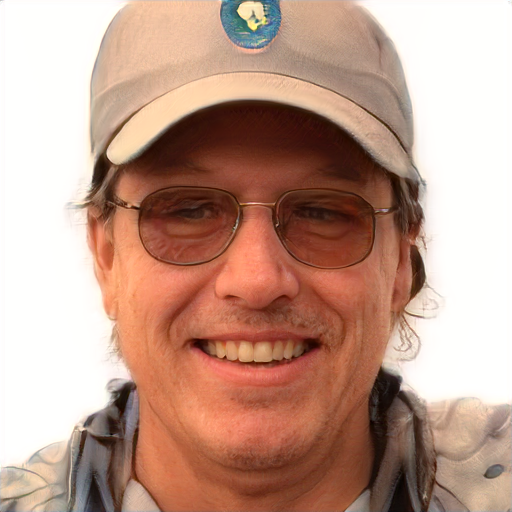} &
    \includegraphics[width=\linewidth]{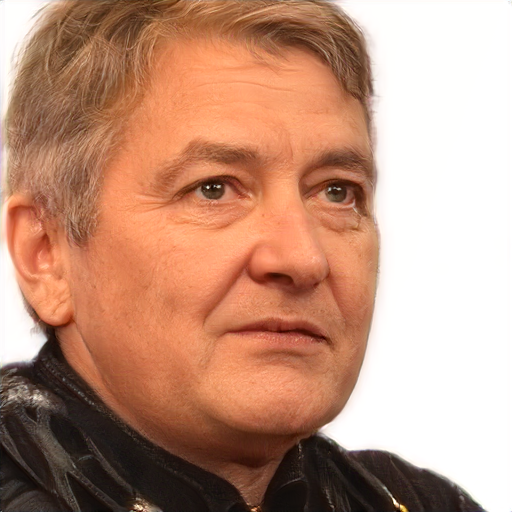} &
    \includegraphics[width=\linewidth]{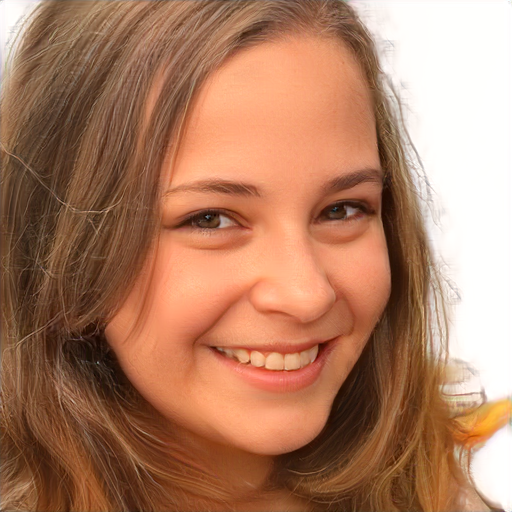} &
    \includegraphics[width=\linewidth]{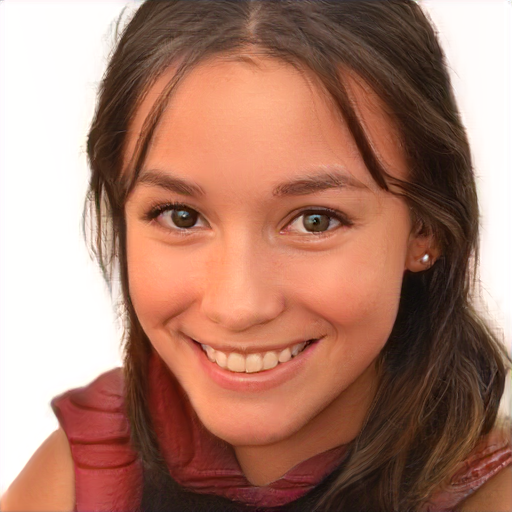} &
    \includegraphics[width=\linewidth]{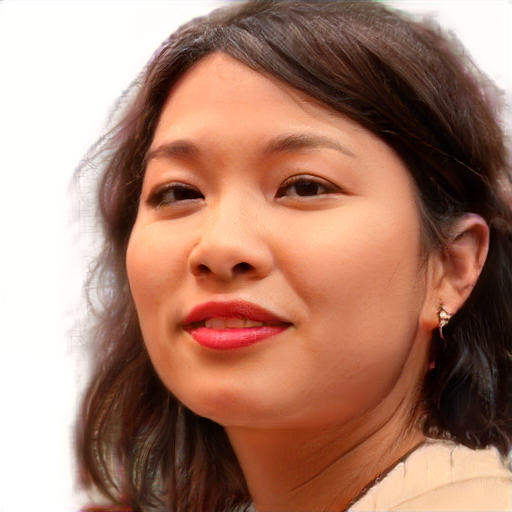}
    \\
    \frame{\includegraphics[height=\linewidth,width=\imgheight,angle=90,origin=c]{figure/paper_uncurated/hdri_library/stream_2k_raw.png}} &
    \includegraphics[width=\linewidth]{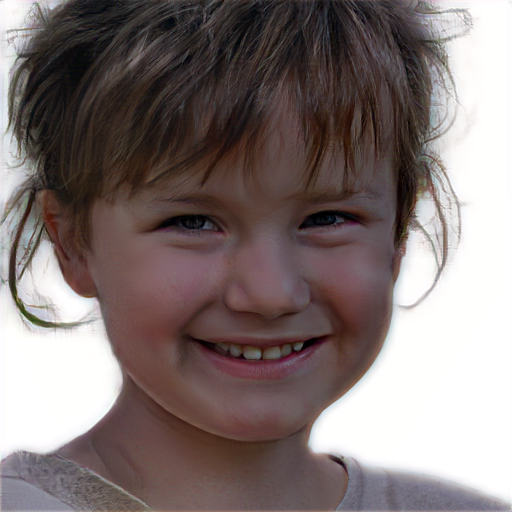} &
    \includegraphics[width=\linewidth]{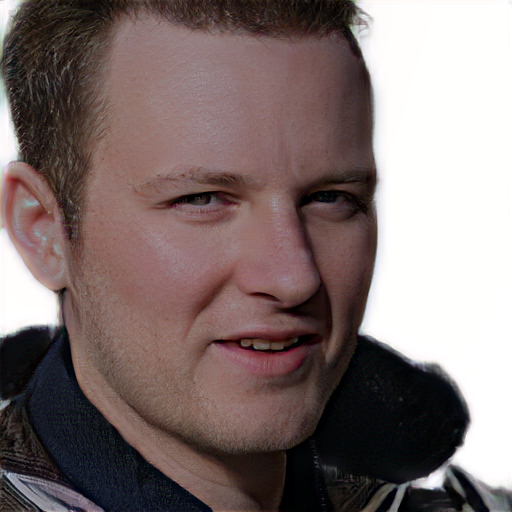} &
    \includegraphics[width=\linewidth]{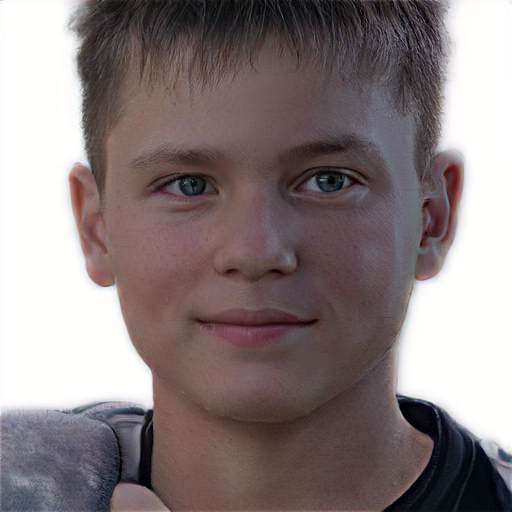} &
    \includegraphics[width=\linewidth]{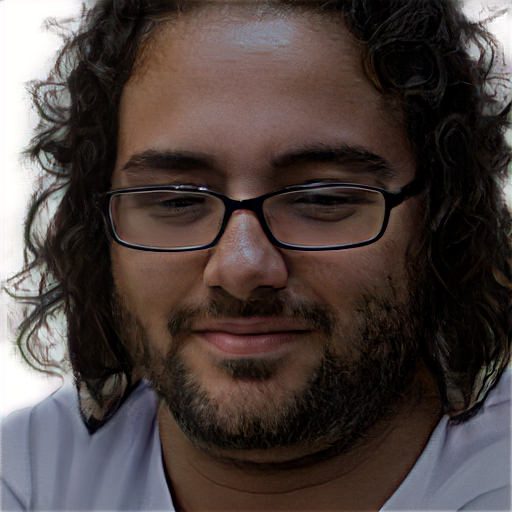} &
    \includegraphics[width=\linewidth]{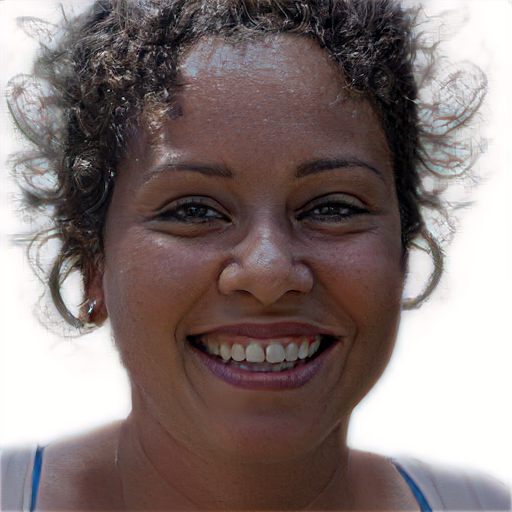} &
    \includegraphics[width=\linewidth]{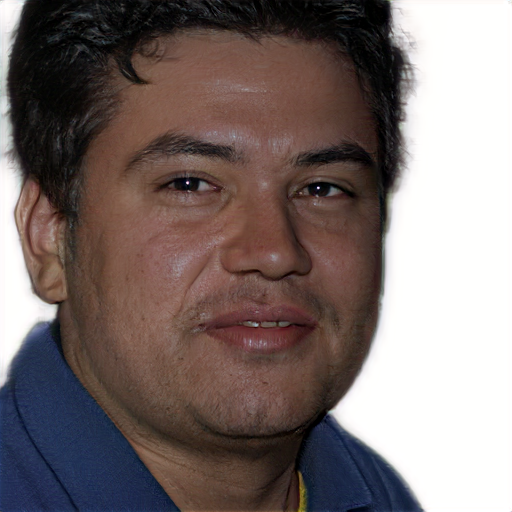} &
    \includegraphics[width=\linewidth]{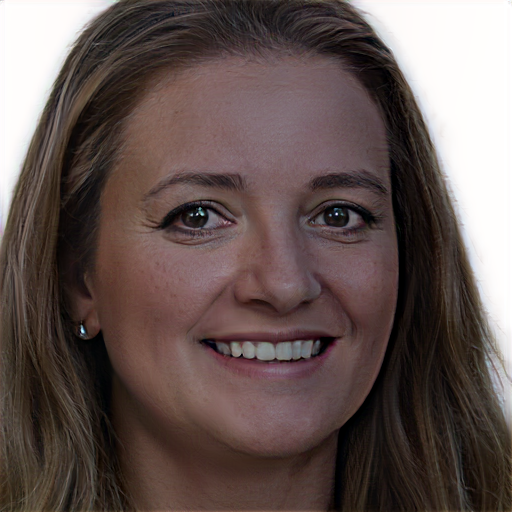}
    \\
    \end{tabular}
    \caption{Uncurated results under random views. Each row corresponds to the same environment lighting. All results are sampled with truncation $\psi=0.7$. As exemplified in the last lighting condition, our method generates more realistic lit result, while VoluxGAN shows a global color shift that does not match the given lighting condition. We hypothesize this is due to the fact that VoluxGAN applies additional 2D convolutional layers to the high-resolution rendered image, introducing physically inaccurate artifacts.}
    \label{fig:uncurated_result}
\end{figure*}
In \cref{fig:uncurated_result} we provide uncurated examples generated using our method, VoluxGAN~\cite{tan2022voluxgan} and ShadeGAN~\cite{pan2021shading}.
We apply truncation~\cite{brock2018large,karras2019style,marchesi2017megapixel} at 0.7.


\end{document}